\PassOptionsToPackage{table}{xcolor}  
\documentclass{article} 
\usepackage{iclr2027_conference,times}

\usepackage{amsmath,amsfonts,bm}

\def\eqref#1{equation~\ref{#1}}

\def\1{\bm{1}}

\DeclareMathAlphabet{\mathsfit}{\encodingdefault}{\sfdefault}{m}{sl}
\SetMathAlphabet{\mathsfit}{bold}{\encodingdefault}{\sfdefault}{bx}{n}

\usepackage{wrapfig}
\usepackage{hyperref}
\usepackage{url}
\usepackage{booktabs}
\usepackage{makecell}
\usepackage{graphicx}
\usepackage{amssymb}
\usepackage[table]{xcolor}
\usepackage{multirow}
\usepackage{tabularx}
\usepackage{tcolorbox}
\tcbuselibrary{skins}
\definecolor{egPrompt}{HTML}{3B5B8C}
\newtcolorbox{promptbox}[2][]{enhanced, colback=egPrompt!4, colframe=egPrompt, colbacktitle=egPrompt, coltitle=white,
  boxrule=0.6pt, arc=1.2mm, left=2mm, right=2mm, top=1.5mm, bottom=1.5mm, toptitle=0.6mm, bottomtitle=0.6mm,
  fonttitle=\sffamily\bfseries\small, title=#2,
  before upper={\ttfamily\footnotesize\frenchspacing\raggedright\setlength{\parindent}{0pt}\setlength{\parskip}{0.6em}}, #1}
\newcommand{\ph}[1]{\textcolor{egPrompt}{\ttfamily\bfseries\{#1\}}}   
\newcommand{\dq}{\char34}                                      
\usepackage{booktabs,graphicx,xcolor,colortbl}

\definecolor{egGemini}{HTML}{E5E9F0}
\definecolor{egQwen35}{HTML}{DFEBFA}
\definecolor{egQwen3VL}{HTML}{DAF1F2}
\definecolor{egQwen25VL}{HTML}{DFF2E2}
\definecolor{egGemma}{HTML}{FCE9D7}
\definecolor{egIntern}{HTML}{EDF0D8}
\definecolor{egGLM}{HTML}{EEE4F7}
\definecolor{egKimi}{HTML}{F9E1E9}

\newcommand{\egfamilyrow}[2]{%
  \rowcolor{#1}\multicolumn{8}{@{}l}{\textit{#2}}\\%
}
\newcommand{\egresultrow}[9]{%
  \rowcolor{#1}#2 & #3 & #4 & #5 & #6 & #7 & #8 & #9\\%
}
\newcommand{\egyes}{\ensuremath{\checkmark}}
\newcommand{\egnr}{\textemdash}

\title{Does Local Video Understanding Transfer Across Encounters? The EgoGears Benchmark}

\iclrfinalcopy
\author{Yuedong Tan$^{1,}$$^{4}$\footnotemark[1] \quad Lei Qi$^{2}$\thanks{Equal contribution.} \quad Yu Liu$^5$ \quad Di Wen$^{3}$ \quad Ruiping Liu$^{3}$ \quad Xiaoye Wang$^{1}$ \\ \textbf{Yufan Chen$^{3}$ \quad Junwei Zheng$^{3}$ \quad Chengzhi Wu$^{3}$ \quad Chen Zhang$^{3}$ \quad Zhihang Chen$^{3}$} \\ \textbf{ Haiwen Sun$^{3}$ \quad Zongwei Wu$^{4}$ \quad Radu Timofte$^{4}$ \quad Danda Pani Paudel$^{1}$ \quad Kunyu Peng$^{3}$\thanks{Corresponding author: kunyu.peng@kit.edu}} \\
  $^1$INSAIT, Sofia University ``St. Kliment Ohridski'' \quad
  $^2$Technical University of Munich \quad \\
  $^3$Karlsruhe Institute of Technology  \quad
  $^4$University of Würzburg \quad
  $^5$Institute of Information Engineering
}

\begin{document}

\maketitle
\begin{abstract}
Embodied systems must make knowledge acquired during one encounter usable in another despite changes in viewpoint, motion, and illumination.
Yet aggregate cross-video accuracy conflates failures of local perception with failures to preserve observation identity, establish correspondence, and compose evidence, obscuring whether local video understanding actually transfers.
We introduce \textbf{EgoGears}, a complementary single- and multi-video benchmark designed to diagnose this transition.
It contains $567$ single-video and $1{,}487$ multi-video questions derived from $126$ human-collected egocentric recordings covering $39$ outdoor routes.
Repeated traversals across movement speeds and lighting conditions ground comparisons in shared physical environments; $531$ questions require alignment across independent recordings.
Single-video questions measure the local visual, spatial, and motion evidence available to a model, while multi-video questions test whether evidence remains bound to the correct observation and can be composed into consistent route relationships.
We report $29$ single-video and $31$ multi-video MLLM configurations across six model families in the main leaderboard.
Among the $20$ configurations evaluated comparably on both splits, every model performs worse on multi-video questions, with a mean decrease of $22.5$ percentage points, and the gap persists when answer format and scoring are held fixed.
The gap is not explained simply by additional videos or recording boundaries. The central bottlenecks are observation--evidence binding and ordered route-state tracking.
EgoGears provides a diagnostic testbed for understanding when, why, and how local video understanding fails to transfer across encounters. 
The code and benchmark are publicly available at \url{https://github.com/lei-qi-233/EgoGears}
\end{abstract}
\section{Introduction}
Recent multimodal large language models (MLLMs) have made rapid progress in video understanding, including long-form question answering, temporal reasoning, and egocentric activity understanding~\citep{EgoVQA, Egocross,peng2026seeing, EgoTempo, EgoMonth}. For outdoor embodied AI, a consequential next step is to make knowledge acquired during one encounter usable in another.
An agent may revisit a junction from the opposite direction, encounter a familiar landmark under different illumination, or compare its current journey with a previously observed route. Successful interpretation requires connecting observations that differ in appearance while preserving distinctions between places, travel directions, and events.
A model must determine what remains the same, what has changed, and which experience supports each conclusion.
\emph{Cross-video comparison thus probes a capability central to embodied intelligence: constructing a consistent understanding of the world from separate, partial encounters.}

Human-collected egocentric videos provide a grounded setting for investigating this capability.
Our study uses recordings of people walking, jogging, and running along outdoor routes, with repeated traversals across movement speeds and lighting conditions.
These recordings capture how bodily movement shapes visual evidence: turns change ego-relative directions, speed changes the temporal spacing between landmarks, and return journeys reverse their order of appearance.
Repeated observations of the same routes anchor these variations to a shared physical environment.
This setting requires models to relate observations through environmental structure while accounting for the circumstances of each encounter.
Its relevance extends to embodied systems that must interpret human route demonstrations, recognize previously visited places, or connect current observations with earlier experience.

The distinctive contribution of our study lies in combining \emph{human egocentric motion, repeated outdoor traversals, and complementary single- and multi-video evaluation}.
Existing benchmarks have advanced local perception, temporal understanding, and reasoning across extended experiences~\citep{EgoVQA, Egocross, EgoTempo,EgoMonth}.
Our focus is the relationship between these levels of competence:
\emph{does understanding individual observations support consistent reasoning across observations?}
The single-video component assesses the local visual, spatial, and motion evidence available to a model.
The multi-video component examines whether the model can retain observation identity, establish correspondence across separate recordings, and compose evidence into relationships among landmarks, directions, and route segments. 
Consider a runner returning at dusk along a route previously recorded in daylight. At an unfamiliar-looking junction, the runner asks: ``Did I pass this junction earlier, and which landmark should appear next on the way back?'' A wearable assistant would need to compare the current observation with the earlier recording, recognize the junction despite changes in illumination and viewing direction, identify the return phase, and infer the reversed landmark sequence.
Recognizing objects in each video independently would leave these relationships unresolved. Together, these evaluations support a more informative assessment of cross-video competence than aggregate accuracy alone.
Cross-video errors may stem from local perception, evidence misattribution, failed correspondence, or reasoning over aligned observations. Evaluating these capabilities jointly helps the community identify where improvement is needed.
We introduce \textbf{EgoGears}, a benchmark that examines whether models can preserve the identity and context of visual evidence when reasoning across outdoor experiences. It comprises $567$ single-video questions and $1{,}487$ multi-video comparison questions derived from $126$ human-collected egocentric recordings covering $39$ outdoor routes and $22.6$ hours.
Repeated traversals across movement speeds and lighting conditions anchor changes in appearance and motion to shared physical environments.
The benchmark connects three evaluation targets: \emph{local understanding}, \emph{evidence integration}, and \emph{cross-recording correspondence}, all in high-motion walking and running footage.
Single-video questions assess objects and attributes, ego-relative spatial relations, ego-motion, trajectories, and temporal events.
Multi-video questions require models to associate evidence with the correct clip, integrate non-contiguous observations, recognize corresponding places and route phases, compare travel directions, identify shared paths, and localize route divergence.
Of these, $531$ questions require alignment across independent recordings.

\begin{table*}[t]
\centering
\caption{
Comparison with representative egocentric and cross-video QA benchmarks. S: single-video QA; M: joint reasoning over multiple videos or clips.
NR: a reliable total video duration was not established from the paper.
}
\label{tab:benchmark_comparison}

\begingroup
\setlength{\tabcolsep}{3pt}
\renewcommand{\arraystretch}{1.2}
\resizebox{\textwidth}{!}{%
\begin{tabular}{@{}llrrccccc@{}}
\toprule
\multirow{2}{*}{Benchmark}
& \multirow{2}{*}{Video source}
& \multirow{2}{*}{\#QA}
& \multirow{2}{*}{\makecell{Video\\hours}}
& \multirow{2}{*}{QA input}
& \multicolumn{4}{c}{Route understanding and acquisition design} \\
\cmidrule(lr){6-9}
& & & & &
\makecell{Route-structure\\QA}
& \makecell{Repeat-path\\protocol}
& \makecell{Multiple\\speed conditions}
& \makecell{Day/night\\counterparts} \\
\midrule

EgoSchema~\citep{EgoSchema}
& Ego4D
& $>5$K
& $>250$
& S
& \egnr & \egnr & \egnr & \egnr \\

EgoTempo~\citep{EgoTempo}
& Ego4D
& 500
& $\sim$4.6
& S
& \egnr & \egnr & \egnr & \egnr \\

EgoCross~\citep{Egocross}
& Five video datasets
& 957
& NR
& S
& \egnr & \egnr & \egnr & \egnr \\

EgoNight~\citep{zhang2026egonight}
& Real + synthetic
& 3,658
& NR
& M
& \egnr & \egnr & \egnr & \egyes \\

EgoExoMem~\citep{liu2026egoexomem}
& \makecell[l]{Ego-Exo4D\\+ LEMMA}
& $\sim$2.6K
& NR
& \makecell{Synchronized\\ego--exo}
& \egnr & \egnr & \egnr & \egnr \\

CrossVid~\citep{CrossVid}
& Six video datasets
& 9,015
& $\sim$318.4
& M
& \egnr & \egnr & \egnr & \egnr \\

\midrule
\rowcolor{blue!7}
\textbf{EgoGears (ours)}
& \textbf{Human outdoor routes}
& \textbf{2,054}
& {22.6}
& \textbf{S + M}
& \egyes & \egyes & \egyes & \egyes \\

\bottomrule
\end{tabular}%
}
\endgroup
\vspace{-8mm}
\end{table*}

The question design makes \emph{the relationship between an observation and its context} central to evaluation.
Questions combine an observation target, a temporal or route-phase condition, and a relation to infer, with answers grounded in specified evidence windows.
Distractors introduce targeted perturbations to visually supported content, including directional reversals, temporal-order changes, route-phase mismatches, and incorrect attribution of landmarks or events.
Resolving these alternatives requires identifying not only which facts are present, but where, when, and to which observation they belong.
Variable-cardinality multi-select questions further require models to recover the complete supported answer set.
Complementary input-structure and capability axes organize questions by how evidence is distributed and what reasoning it demands; five input-structure units span evidence within one recording through correspondence across recordings.
Together with the single-video assessment, this design supports investigating whether difficulties arise in recovering local evidence or in preserving and composing it across observations.
Deterministic option shuffling, metadata anonymization, text-only shortcut screening, video-grounded verification, and human review strengthen the connection between benchmark performance and visual evidence.

We report $29$ single-video and $31$ multi-video MLLM configurations spanning six model families in the main leaderboard.
Our evaluation reveals persistent difficulties with clip--evidence association and ordered route reasoning, with the strongest overall multi-video configuration reaching only $33.3\%$ on multi-step turning over whole-video evidence (Table~\ref{tab:multi_capability}).
Across the eight matched single-video comparisons in the main leaderboard, enabling explicit reasoning improves every configuration by a median of $7.8$ percentage points, but the gains are substantially larger for ego-relative spatial relations than for trajectory-grounded questions ($14.0$ versus $3.7$ points), suggesting that stronger evidence verification does not translate uniformly into better understanding of movement and route structure.
Our contributions are threefold:
\begin{itemize}
    \item We formulate cross-video understanding for outdoor embodied AI around evidence binding, correspondence, and spatiotemporal composition. Joint single- and multi-video evaluation examines whether local understanding supports consistent reasoning across encounters.

    \item We introduce \textbf{EgoGears}: $567$ single-video and $1{,}487$ multi-video questions from $126$ human-collected egocentric recordings across $39$ outdoor routes, including $531$ questions requiring correspondence across independent recordings. Repeated traversals, relational distractors, and exact-set scoring probe evidence consistency across changes in movement speed and illumination.

    \item We report $29$ single-video and $31$ multi-video configurations across six MLLM families, revealing bottlenecks in evidence binding and ordered route reasoning. Across eight matched single-video comparisons, reasoning improves every configuration (median gain: $7.8$ percentage points), with smaller gains for trajectory understanding than ego-relative spatial reasoning.
\end{itemize}

\section{Related Work}
\label{sec:related}

\noindent\textbf{Egocentric Video QA Benchmarks.}
Egocentric video understanding has progressed from action understanding (EgoVQA~\citep{EgoVQA} and EgoTaskQA~\citep{EgoTaskQA}) to long-video temporal memory (EgoTempo~\citep{EgoTempo}) and object-interaction topology (AMEGO~\citep{AMEGO}), yet route-level spatial reasoning under highly dynamic outdoor motion remains underexplored. Spatial QA benchmarks, including NuScenes-QA~\citep{NuScenes}, SpatialBench~\citep{Spatialbench}, and MMSI-Video-Bench~\citep{MMSI}, incorporate outdoor scenes or 3D geometry but primarily rely on vehicle trajectories, third-person views, or static scenes. EgoCross~\citep{Egocross} covers extreme sports but remains limited to single clips. Regarding input scope, EgoThink~\citep{Egothink} uses single video sources, while Ego-Exo4D~\citep{exo4d} and EgoExoMem~\citep{liu2026egoexomem} support synchronized multi-view reasoning. EgoMonth~\citep{EgoMonth} extends reasoning across videos and days but mainly targets routine activities. However, they do not directly test whether knowledge acquired in one encounter remains spatially and temporally coherent when the same physical environment is revisited under different viewpoints, directions, speeds, or illumination.
EgoGears targets this gap by pairing single-video understanding with cross-recording correspondence over repeated outdoor traversals.


\noindent\textbf{Multimodal Large Language Models.}
MLLMs increasingly rely on key-frame selection and token compression to process long, highly dynamic videos~\citep{shi2025slow,jiang2025storm,peng2026seeing,wen2026provia}. General-purpose open-source models, including InternVL~\citep{zhu2025internvl3}, MiniCPM~\citep{hu2024minicpm}, and Qwen-Omni~\citep{xu2025qwen3}, alongside egocentric models such as EgoVLPv2~\citep{pramanick2023egovlpv2} and EgoGPT~\citep{yang2025egolife}, have advanced dynamic visual representation. Qwen3-VL~\citep{bai2025qwen3} further supports ultra-long sequences through Interleaved-MRoPE and explicit text-based temporal alignment. Commercial Gemini models, including Gemini 3.1 Pro~\citep{deepmind2026gemini31pro}, provide strong general-purpose reasoning baselines and are included in our evaluation. Recent MLLMs nevertheless struggle with physical spatiotemporal and multi-video reasoning~\citep{MMSI}, and most open-source MLLMs lack comprehensive cross-video reasoning training. Frame sampling and token compression can discard spatial landmarks essential for continuous state tracking~\citep{shi2025mavors,shi2025slow}, while outdoor lighting changes exacerbate hallucinations and reliance on language priors~\citep{wu2025lanp,zhuang2026mitigating}. Under degraded visibility, misalignment between these priors and visual evidence, or indiscriminate suppression of the priors, can further disrupt semantic consistency and state tracking.
We therefore report 29 single-video and 31 multi-video configurations in the main leaderboard to assess highly dynamic state tracking, environmental robustness, and asynchronous cross-video reasoning. For configurations evaluated in comparable modes, multi-video accuracy is consistently lower than single-video accuracy, also when answer format and scoring are held fixed. The results motivate closer analysis of evidence binding and ordered route reasoning without attributing the observed difference to the number of videos alone.

\begin{figure*}[t]
    \centering
    \includegraphics[width=\textwidth]{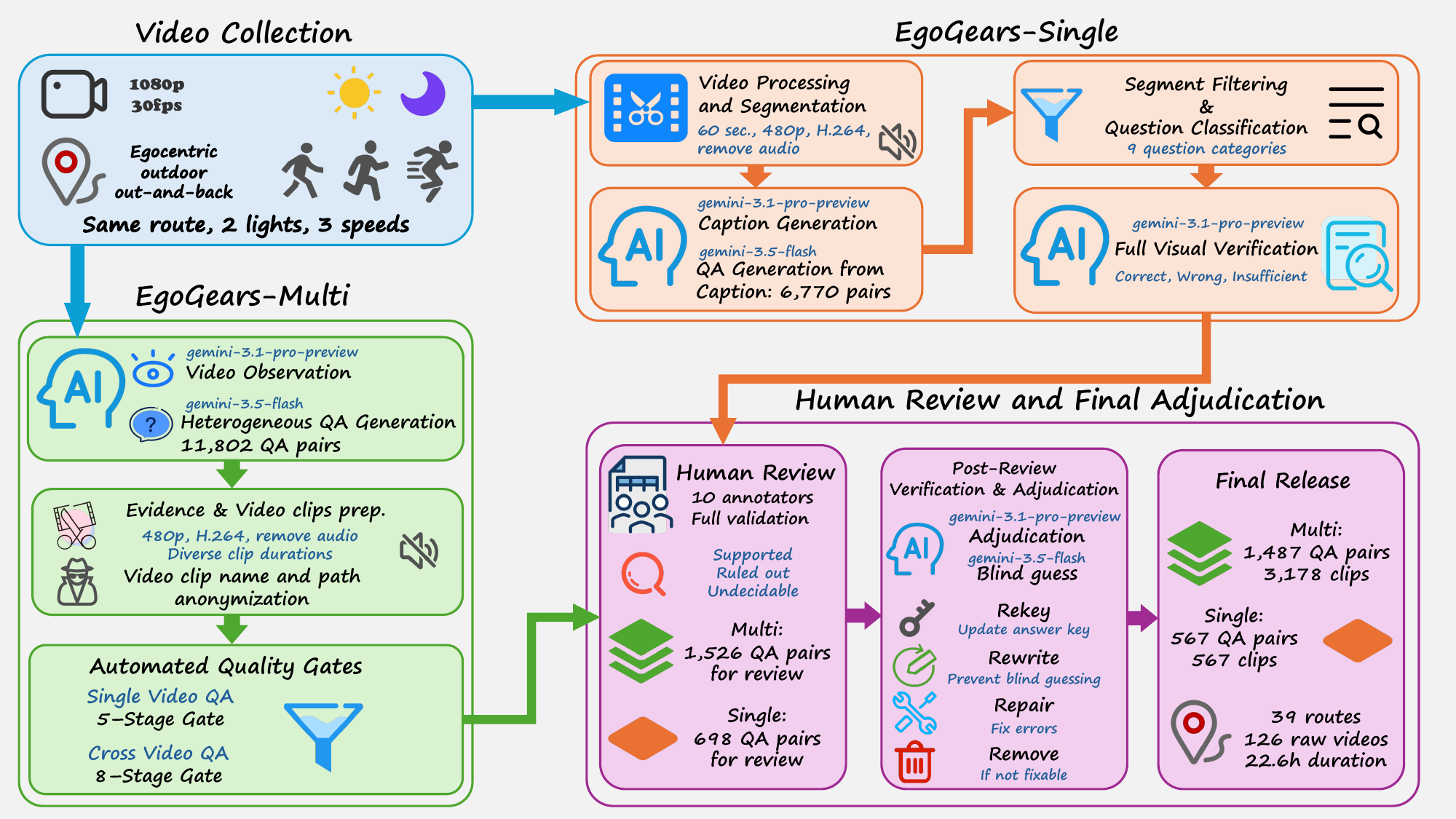}
    \caption{Overview of the EgoGears collection and annotation pipeline. Controlled outdoor route recordings feed separate candidate-generation paths for EgoGears-Single and EgoGears-Multi. Automated gates screen structure, leakage, guessability, and visual support before human review. Independent post-review verification and adjudication then retain, revise, repair, or remove questions to produce the final paired benchmark.}
    \label{fig:annotation_pipeline}
\end{figure*}

\section{EgoGears Dataset}

Figure~\ref{fig:annotation_pipeline} summarizes how EgoGears is built. We record $126$ egocentric videos over $39$ outdoor routes at three movement speeds (slow walking, fast walking, and running) and two lighting conditions (day and night); coverage varies by route (Section~\ref{sec:dataset}). EgoGears-Single and EgoGears-Multi are generated along separate paths and screened by automated quality gates. The candidates then pass through human review and post-review adjudication, which yields $567$ single-video and $1{,}487$ multi-video QA pairs. We first describe the QA design principles shared by both subsets (Sec.~\ref{sec:design}).

\subsection{QA Pairs Design}
\label{sec:design}

\noindent\textbf{Design Objective and Task Scope.} This study targets first-person outdoor high-dynamic video understanding, primarily focusing on walking, fast walking, and running scenarios. These videos inherently contain continuous camera displacement, viewpoint changes, motion-induced instability, and dynamic external entities. Under varying controlled conditions (e.g., changes in illumination and movement speed), the same route may exhibit different visual characteristics. Unlike traditional video QA that focuses on static scene summarization or isolated frame recognition, our central objective is to assess whether models can form a local, continuous-temporal, and cross-clip understanding of first-person routes. Questions require models to connect visual entities with their temporal position, spatial relation, route phase, and motion state. The model must determine not only \textit{what} appears, but \textit{when} it appears, its relative position to the walking path, and its order in the route structure, demonstrating a structured spatiotemporal representation of the dynamic environment.

\noindent\textbf{Principles of Question Construction.} Every question must be traceable to defined video evidence windows. A question generally encodes three hierarchical levels of information: \underline{Observation Target}, indicating specific entities such as buildings, vehicles, or pedestrians; \underline{Observation Condition}, defining contextual constraints like specific time intervals, points before or after a turn, or outbound versus return legs; and \underline{Inferred Relation}, specifying the reasoning objective, such as presence, relative direction, sequence, or cross-video correspondence. This structure ensures the task measures visually grounded spatiotemporal reasoning rather than text-based memorization. To achieve this, visual facts are transformed into discriminative relations to prevent models from relying on coarse scene-level categorization. For instance, rather than asking if an object exists, questions ask which objects appear \textit{after a right turn}, or the sequence of passed landmarks. Spatial relations emphasize relative metrics (left/right, ahead/behind) based on the camera wearer's trajectory, while dynamic entities are distinguished by their behavior (e.g., approaching, overtaking, crossing). Furthermore, questions scale in difficulty by utilizing single continuous clips, multiple non-contiguous windows, or complete route stages. This gradient tests the model's ability to transition from local observation to long-temporal reasoning. Crucially, the provided evidence windows must jointly cover the asked relation; questions requiring unobservable context are excluded from deterministic evaluation.

\noindent\textbf{Option and Answer Design.} EgoGears employs structured, letter-based closed-set options. For multiple-choice formats, a response is strictly evaluated based on the complete set of correct options. This design forces the model to perform both evidence-support and evidence-exclusion judgments simultaneously, ensuring it identifies all relevant visual evidence rather than stopping at one salient cue. Distractors are engineered as near-miss statements that apply controlled perturbations to entities genuinely present in the video, preventing models from guessing answers based on noun-matching or surface-level syntactic cues. Key perturbation strategies include spatial and directional reversals (e.g., swapping left/right or reversing turn directions), temporal and phase mismatches (e.g., reversing event order or incorrectly applying an outbound observation to a return leg), entity state confusion (e.g., confusing a parked object with a moving one, or altering the direction of motion), and subject-relation reassignment (e.g., pairing a legitimate landmark with the wrong building or action). All options are normalized for length, syntax, and information density. Ultimately, answer judgment is deeply grounded in visual evidence, explicitly distinguishing between a definitively \textit{incorrect} statement and an \textit{indeterminate} one. If a distractor incorrectly claims a spatial or temporal state, it is marked incorrect. However, if the video footage simply lacks the evidence to verify an option (e.g., a landmark is absent from a specific short clip but may exist in the broader unshown route), the option is not forced into the incorrect category, ensuring the evaluation remains strictly deterministic.

The EgoGears QA design establishes a multi-granular evaluation framework centered on visual-evidence-driven first-person route understanding. By combining structured spatiotemporal relations, closed-set answer evaluations, and visually perturbed distractors, the design transcends basic video captioning. It robustly challenges models to decode the complex interplay of route trajectory, observer viewpoint, continuous time, and external dynamics.
\subsection{Collection Methodology}
\label{sec:methodology}

\noindent\textbf{Human-collected route observations.}
We collect human egocentric recordings of outdoor routes across movement speeds and day/night illumination. Routes include distinctive landmarks and diverse turn structures while minimizing incidental pedestrian capture. Repeated and out-and-back traversals provide observations of shared physical structure under changes in viewpoint, travel direction, motion, and lighting. Condition coverage varies across routes, as reported in Section~\ref{sec:dataset}.

\noindent\textbf{Evidence-grounded question generation.}
MLLMs extract structured observations of landmarks, turns, route phases, spatial relations, and events to generate candidate questions. EgoGears-Single uses continuous $60$-second clips, including shorter tail clips, to assess local understanding. EgoGears-Multi combines whole-recording or segment-level observations and, where required, independent recordings to assess route-phase alignment, landmark correspondence, event attribution, and route divergence.
Questions associate a visual target with an observation or route-phase condition and require inferring a spatial, temporal, or cross-observation relation. Relational distractors perturb direction, ordering, route phase, entity state, or evidence attribution.

\noindent\textbf{Preserving observation boundaries.}
Each question is linked to its source recordings and evidence windows. Non-contiguous excerpts and clips from different recordings remain separate inputs, preserving the observation boundaries needed for evidence binding. Evaluation clips use 480p H.264 video without audio, and anonymized identifiers conceal recording metadata.

\noindent\textbf{Verification and review.}
Text-only screening flags potentially guessable questions, while video-grounded verification evaluates options as supported, ruled\_out, or undecidable. Multi-clip and cross-video candidates additionally undergo checks for comparison requirements, evidence availability, duplication, and metadata leakage; deterministic shuffling controls option-position and clip-order cues.
Human reviewers assess question clarity and option validity. Multi-video reviewers additionally construct separate clip timelines and assess correspondence using stable physical structure rather than transient appearance.
Adjudication combines reviewer judgments with further visual and text-only checks; repaired questions are rechecked, and unresolved cases are removed. The pipeline yields $567$ single-video and $1{,}487$ multi-video questions. Acquisition settings, model roles, screening procedures, and review coverage are detailed in Appendix~\ref{app:collection_details}.

\begin{figure*}[t]
    \centering
    \includegraphics[width=\textwidth]{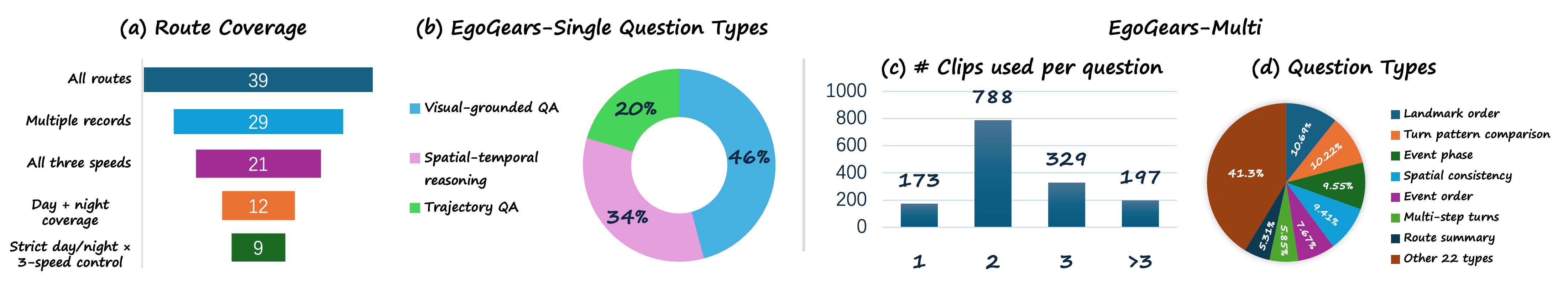}
\caption{\textbf{EgoGears statistics.}
\textbf{(a)} Route coverage: of the 39 routes, 29 have multiple recordings, 21 cover all three movement speeds, 12 have both day and night recordings that together cover all three speeds, and 9 are fully crossed, with one recording under each of the six speed $\times$ lighting conditions.
\textbf{(b)} EgoGears-Single question types ($N=567$).
\textbf{(c)} Number of clips per EgoGears-Multi question ($N=1{,}487$).
\textbf{(d)} EgoGears-Multi question types; the seven most frequent of the 29 fine-grained types are shown individually, and the remaining 22 are grouped.}
\label{fig:dataset_stats}
\vspace{-3mm}
\end{figure*}

\subsection{Dataset Composition}
\label{sec:dataset}

EgoGears evaluates local video understanding and cross-video reasoning using human-collected outdoor egocentric recordings spanning different movement speeds and lighting conditions. Figure~\ref{fig:dataset_stats} summarizes route coverage and question distributions.

\noindent\textbf{Source recordings and route coverage.}
The collection contains $126$ recordings across $39$ routes, totaling approximately $22.6$ hours. It includes $88$ daytime and $38$ nighttime recordings, with $52$ slow-walking, $32$ fast-walking, and $42$ running videos; $54$ recordings follow out-and-back trajectories.
Of the $39$ routes, $29$ have multiple recordings, $35$ include daytime observations, and $16$ include nighttime observations. Movement and lighting coverage varies by route: $21$ routes cover all three movement categories, while $12$ include both lighting conditions and collectively cover all movement categories. Among these, $9$ contain all six movement--lighting combinations, enabling comparisons of each movement category under both day and night conditions.

\noindent\textbf{Clip preparation.}
Evaluation clips are standardized to 480p H.264 with audio removed. Unique identifiers link each question to its visual evidence: one clip file for EgoGears-Single and a folder of question-specific clips for EgoGears-Multi.

\noindent\textbf{EgoGears-Multi.}
This subset contains $1{,}487$ verified QA pairs with supporting evidence and explanations. Five input structures distinguish how evidence is distributed. Whole-video questions and questions using multiple clips from the same recording account for $64.29\%$ of the subset. The remaining $531$ questions ($35.71\%$) require correspondence across independent recordings through anonymous multi-clip inputs, explicit cross-video reasoning, or cross-recording comparisons.
The subset covers $29$ fine-grained question types, including landmark order, turn-pattern comparison, event-to-route-phase alignment, spatial consistency, route summarization, and landmark revisits. These questions test whether models associate evidence with the correct observation and compose it into consistent spatial and temporal relations.

\noindent\textbf{EgoGears-Single.}
This subset contains $567$ verified QA pairs, each grounded in one continuous clip: $540$ question inputs are $60$ seconds long, and $27$ are shorter tail clips. It assesses local understanding without requiring cross-clip correspondence.
Questions span three core types: visually grounded QA ($45.86\%$), spatiotemporal reasoning ($33.69\%$), and trajectory QA ($20.46\%$). Nine fine-grained categories further characterize capabilities, including object and attribute recognition, ego-relative spatial reasoning, trajectory understanding, ego-motion recognition, scene understanding, and OCR.

\noindent\textbf{Answer format and shortcut controls.}
Both subsets vary the number of candidate options and correct answers. EgoGears-Multi provides $6$ or $8$ options, with $1$--$7$ correct answers. EgoGears-Single provides $4$--$8$ options and combines single-answer ($57.32\%$) and multiple-answer ($42.68\%$) questions; the latter contain $2$--$5$ correct answers. Exact-set scoring requires selecting all and only the correct options. Metadata are anonymized to limit nonvisual shortcuts. Ten annotators cross-checked all questions (about $2{,}100$ hours); Gemini API usage cost about USD $30{,}000$.

\begin{table*}[t!]
\vspace{-5mm}
\centering
\begin{minipage}[t]{0.49\textwidth}
\vspace{0pt}
\centering
\caption{\textbf{Single-video QA} ($N=567$). Exact-set accuracy (\%).}
\label{tab:single_results}
\begingroup
\fontsize{7.2}{8.2}\selectfont
\setlength{\tabcolsep}{1.8pt}
\renewcommand{\arraystretch}{1.22}
\resizebox{\linewidth}{!}{%
\begin{tabular}{@{}lcrrrrrr@{}}
\toprule
Model & Mode & \textbf{All} & Obj. & Sp. & Tr. & SC & MS \\
\midrule
\egfamilyrow{egQwen35}{Qwen3.5}
\egresultrow{egQwen35}{122B-A10B}{T}{\textbf{68.6}}{72.2}{\textbf{77.2}}{45.8}{68.9}{\textbf{68.2}}
\egresultrow{egQwen35}{}{D}{59.1}{65.6}{57.3}{40.6}{62.2}{55.0}
\egresultrow{egQwen35}{35B-A3B}{T}{66.5}{68.4}{67.8}{51.0}{67.7}{64.9}
\egresultrow{egQwen35}{}{D}{60.8}{66.5}{60.2}{42.7}{63.4}{57.4}
\egresultrow{egQwen35}{27B}{T}{\textbf{68.6}}{68.9}{75.4}{\textbf{53.1}}{\textbf{69.5}}{67.4}
\egresultrow{egQwen35}{}{D}{62.6}{69.4}{62.6}{45.8}{65.5}{58.7}
\egresultrow{egQwen35}{9B}{T}{58.6}{62.7}{58.5}{43.8}{59.7}{57.0}
\egresultrow{egQwen35}{}{D}{48.5}{55.5}{43.3}{35.4}{52.6}{43.0}

\addlinespace[2pt]
\egfamilyrow{egQwen3VL}{Qwen3-VL}
\egresultrow{egQwen3VL}{235B-A22B}{T}{63.3}{68.4}{66.7}{36.5}{62.8}{64.0}
\egresultrow{egQwen3VL}{}{D}{52.9}{64.1}{38.0}{39.6}{60.6}{42.6}
\egresultrow{egQwen3VL}{30B-A3B}{T}{56.1}{65.1}{49.1}{42.7}{58.2}{53.3}
\egresultrow{egQwen3VL}{8B}{T}{52.6}{60.3}{48.5}{35.4}{52.6}{52.5}
\egresultrow{egQwen3VL}{}{D}{40.7}{52.2}{19.3}{38.5}{54.5}{22.3}

\addlinespace[2pt]
\egfamilyrow{egQwen25VL}{Qwen2.5-VL}
\egresultrow{egQwen25VL}{72B}{D}{51.7}{61.7}{38.0}{40.6}{60.3}{40.1}
\egresultrow{egQwen25VL}{7B}{D}{22.4}{25.4}{11.7}{20.8}{35.1}{5.4}

\addlinespace[2pt]
\egfamilyrow{egGemma}{Gemma}
\egresultrow{egGemma}{4-31B}{T}{68.4}{\textbf{74.2}}{70.2}{47.9}{\textbf{69.5}}{66.9}
\egresultrow{egGemma}{}{D}{62.3}{68.4}{62.6}{45.8}{64.3}{59.5}
\egresultrow{egGemma}{4-26B-A4B}{D}{49.0}{61.2}{35.1}{37.5}{55.1}{40.9}
\egresultrow{egGemma}{3n-E4B}{D}{19.8}{15.8}{9.4}{34.4}{28.9}{7.4}

\addlinespace[2pt]
\egfamilyrow{egIntern}{InternVL3.5}
\egresultrow{egIntern}{241B-A28B}{D}{45.7}{53.1}{26.9}{51.0}{58.8}{28.1}
\egresultrow{egIntern}{38B}{D}{27.5}{34.4}{17.0}{21.9}{35.4}{16.9}
\egresultrow{egIntern}{30B-A3B}{D}{27.3}{27.3}{17.5}{34.4}{39.4}{11.2}
\egresultrow{egIntern}{20B-A4B}{D}{39.9}{48.8}{22.2}{39.6}{55.4}{19.0}
\egresultrow{egIntern}{14B}{D}{36.3}{36.8}{24.0}{39.6}{51.1}{16.5}
\egresultrow{egIntern}{8B}{D}{27.0}{28.2}{19.3}{24.0}{37.2}{13.2}

\addlinespace[2pt]
\egfamilyrow{egGLM}{GLM}
\egresultrow{egGLM}{4.6V}{T}{51.1}{57.4}{48.5}{32.3}{51.4}{50.8}
\egresultrow{egGLM}{}{D}{45.9}{53.6}{35.7}{33.3}{51.1}{38.8}
\egresultrow{egGLM}{4.1V-9B}{T}{41.3}{51.7}{28.1}{28.1}{47.1}{33.5}

\addlinespace[2pt]
\egfamilyrow{egKimi}{Kimi}
\egresultrow{egKimi}{VL-A3B-2506}{T}{45.7}{53.6}{32.2}{38.5}{55.4}{32.6}
\bottomrule
\end{tabular}%
}
\endgroup
\end{minipage}\hfill%
\begin{minipage}[t]{0.49\textwidth}
\vspace{0pt}
\centering
\caption{\textbf{Multi-video QA} ($N=1{,}487$). Exact-set accuracy (\%).}
\label{tab:multi_results}
\begingroup
\fontsize{7.2}{8.2}\selectfont
\setlength{\tabcolsep}{1.8pt}
\renewcommand{\arraystretch}{1.14}
\resizebox{\linewidth}{!}{%
\begin{tabular}{@{}lcrrrrrr@{}}
\toprule
Model & Mode & \textbf{All} & W & MC & AN & CV & CR \\
\midrule
\egfamilyrow{egGemini}{Gemini (API)}
\egresultrow{egGemini}{3.8-Flash}{API}{\textbf{78.3}}{\textbf{69.2}}{\textbf{88.8}}{\textbf{84.4}}{70.5}{\textbf{94.6}}
\egresultrow{egGemini}{3.1-Pro}{API}{66.5}{60.1}{75.0}{66.3}{\textbf{71.2}}{74.4}
\egresultrow{egGemini}{2.5-Pro}{API}{56.4}{51.5}{65.1}{54.1}{53.0}{69.0}

\midrule
\egfamilyrow{egQwen35}{Qwen3.5}
\egresultrow{egQwen35}{122B-A10B}{T}{37.8}{42.0}{42.8}{35.2}{15.9}{32.6}
\egresultrow{egQwen35}{}{D}{16.3}{25.2}{14.1}{6.3}{6.8}{7.8}
\egresultrow{egQwen35}{35B-A3B}{T}{37.9}{41.4}{41.8}{34.1}{24.2}{32.6}
\egresultrow{egQwen35}{}{D}{15.1}{25.0}{11.2}{5.2}{6.1}{4.7}
\egresultrow{egQwen35}{27B}{T}{40.8}{40.8}{46.7}{43.0}{22.7}{40.3}
\egresultrow{egQwen35}{}{D}{16.0}{23.9}{10.9}{7.8}{9.1}{12.4}
\egresultrow{egQwen35}{9B}{T}{21.3}{27.6}{18.1}{12.2}{15.9}{21.7}
\egresultrow{egQwen35}{}{D}{10.9}{18.6}{4.9}{2.2}{6.8}{8.5}

\addlinespace[2pt]
\egfamilyrow{egQwen3VL}{Qwen3-VL}
\egresultrow{egQwen3VL}{235B-A22B}{T}{46.0}{52.1}{45.1}{37.4}{37.1}{44.2}
\egresultrow{egQwen3VL}{}{D}{28.6}{38.2}{16.1}{14.4}{40.2}{27.1}
\egresultrow{egQwen3VL}{30B-A3B}{T}{26.2}{35.9}{17.8}{18.5}{25.8}{13.2}
\egresultrow{egQwen3VL}{8B}{D}{16.9}{27.3}{3.9}{7.4}{25.8}{5.4}

\addlinespace[2pt]
\egfamilyrow{egQwen25VL}{Qwen2.5-VL}
\egresultrow{egQwen25VL}{72B}{D}{26.4}{35.6}{11.8}{13.3}{41.7}{25.6}
\egresultrow{egQwen25VL}{7B}{D}{9.8}{16.6}{2.3}{3.3}{13.6}{3.1}

\addlinespace[2pt]
\egfamilyrow{egGemma}{Gemma}
\egresultrow{egGemma}{4-31B}{T}{56.2}{54.6}{60.5}{50.0}{61.4}{61.2}
\egresultrow{egGemma}{}{D}{33.2}{40.5}{25.0}{24.1}{32.6}{34.9}
\egresultrow{egGemma}{4-26B-A4B}{D}{22.0}{32.4}{8.2}{12.2}{29.5}{14.7}
\egresultrow{egGemma}{3n-E4B}{D}{9.0}{13.5}{2.3}{4.8}{14.4}{5.4}

\addlinespace[2pt]
\egfamilyrow{egIntern}{InternVL3.5}
\egresultrow{egIntern}{241B-A28B}{D}{24.2}{34.4}{12.2}{10.4}{40.2}{14.0}
\egresultrow{egIntern}{38B}{D}{18.1}{26.1}{6.9}{8.5}{30.3}{11.6}
\egresultrow{egIntern}{30B-A3B}{D}{13.4}{20.6}{3.9}{4.8}{23.5}{7.0}
\egresultrow{egIntern}{20B-A4B}{D}{13.9}{23.8}{3.3}{4.8}{17.4}{4.7}
\egresultrow{egIntern}{14B}{D}{16.5}{27.0}{3.0}{7.0}{26.5}{4.7}
\egresultrow{egIntern}{8B}{D}{10.4}{15.8}{4.6}{5.6}{15.2}{2.3}

\addlinespace[2pt]
\egfamilyrow{egGLM}{GLM}
\egresultrow{egGLM}{4.6V}{T}{39.9}{35.4}{47.0}{37.4}{47.0}{43.4}
\egresultrow{egGLM}{}{D}{36.9}{35.4}{38.8}{31.5}{53.0}{34.1}
\egresultrow{egGLM}{4.1V-9B}{T}{13.4}{13.3}{9.2}{9.6}{29.5}{15.5}

\addlinespace[2pt]
\egfamilyrow{egKimi}{Kimi}
\egresultrow{egKimi}{VL-A3B-2506}{T}{19.5}{28.7}{12.2}{7.8}{27.3}{7.0}
\bottomrule
\end{tabular}%
}
\endgroup
\end{minipage}

\vspace{4pt}
\begin{minipage}{\textwidth}
\fontsize{7}{8.2}\selectfont
\textbf{Notes.}
Background colors identify model families; variants inherit the italicized
family name above them. Bold indicates the best score in each column.
T = Thinking; D = Direct;
Obj. = objects/attributes; Sp. = ego-relative spatial relations;
Tr. = trajectory; SC = single-choice; MS = multi-select.
W = whole-video; MC = multi-caption multi-clip; AN = anonymous multi-clip;
CV = cross-video; CR = cross-recording.

\textbf{Model references.}
Qwen3.5~\citep{qwen2026qwen35}; Qwen3-VL~\citep{bai2025qwen3};
Qwen2.5-VL~\citep{bai2025qwen25vl}; Gemma-4 and Gemma-3n~\citep{gemmateam2026gemma4,gemmateam2025gemma3};
InternVL3.5~\citep{wang2025internvl35}; GLM-4.6V and GLM-4.1V~\citep{zai2026glm46v,zai2025glm41v};
Kimi-VL~\citep{kimiteam2025kimivl}; Gemini-3.8-Flash, Gemini-3.1-Pro, and
Gemini-2.5-Pro~\citep{deepmind2026gemini38flash,deepmind2026gemini31pro,comanici2025gemini25}.
\end{minipage}
\end{table*}

\section{Experiments}
\label{sec:experiments}

Our experiments examine the transition from local video understanding to reliable comparison across encounters, focusing on evidence attribution, ordered route understanding, and the benefits and limitations of explicit reasoning.

\noindent\textbf{Experimental setup.}
\label{sec:exp_setup}
We evaluate $29$ single-video and $31$ multi-video configurations across six MLLM families on EgoGears.
We use deterministically shuffled options, temperature-zero decoding, and a common structured-answer instruction.
The primary metric is fixed-denominator exact-set accuracy: invalid and unfinished outputs count as incorrect.
\begin{wrapfigure}{r}{0.5\textwidth}
    \vspace{-12pt}
    \centering
    \includegraphics[width=\linewidth]{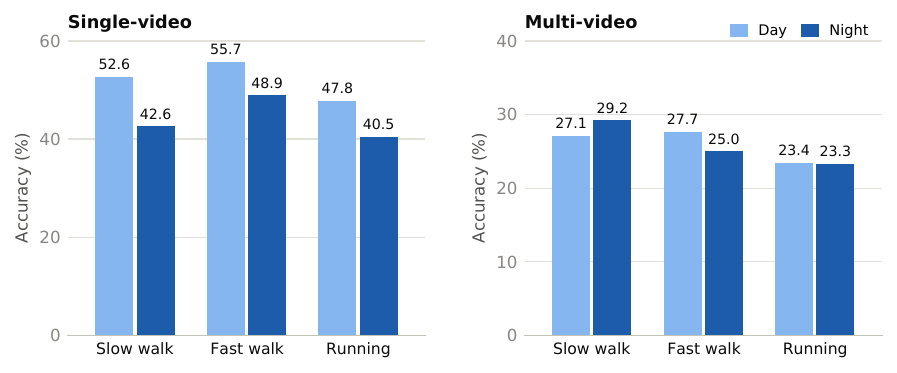}
    \vspace{-16pt}
    \caption{Mean exact-set accuracy on the 9 fully crossed routes (29 single-video and 31 multi-video configurations).}
    \label{fig:condition-bar}
    \vspace{-10pt}
\end{wrapfigure}
\noindent\textbf{Local accuracy leaves cross-video reliability unresolved.}
Tables~\ref{tab:single_results} and~\ref{tab:multi_results} reveal substantially different performance profiles across the two subsets.
Qwen3.5-27B-Thinking and Gemma-4-31B-Thinking achieve nearly identical single-video accuracy ($68.6\%$ and $68.4\%$), but their multi-video scores differ by $15.4$ percentage points ($40.8\%$ versus $56.2\%$).
Both splits are drawn from the same recordings, and the gap is not an artifact of answer format or scoring: all $20$ configurations evaluated on both splits remain lower on multi-video questions among single-choice questions (mean $14.6$ points), among multi-select questions ($20.4$), and under partial credit ($13.3$; Appendix~\ref{app:split_gap}). Local-video evaluation alone is therefore insufficient to assess reasoning across encounters.

\noindent\textbf{Recording identity alone does not determine difficulty.}
Evidence from independent recordings is not uniformly harder than evidence from one recording.
Multi-caption questions draw all clips from one recording, yet $19$ of the $31$ configurations score higher on explicit cross-video questions; InternVL3.5-241B, for example, obtains $40.2\%$ versus $12.2\%$.
Because the units differ in answer format (random exact accuracy $12.6\%$ versus $1.8\%$), we also compare chance-normalized accuracy $\kappa$ (Appendix~\ref{app:metrics}).
The ordering still depends on the model: InternVL3.5, Qwen2.5-VL, and the direct-answer Qwen3-VL models remain stronger on cross-video questions (e.g., $\kappa=31.6$ versus $10.6$ for InternVL3.5-241B), whereas Gemini and every Qwen3.5 configuration are stronger on multi-caption questions.
Which input structure is harder is therefore model-specific rather than determined by whether clips come from different recordings.

\noindent\textbf{Effect of speed and lighting.}
On the 9 routes recorded under all six speed $\times$ lighting conditions, every comparison holds the route and wearer fixed (Fig.~\ref{fig:condition-bar}; 95\% CIs in brackets). Darkness mainly affects perception: night-time changes single-video accuracy by $-8.0$\,pp [$-13.9$, $-1.3$], with 28 of 29 models worse, but multi-video accuracy by only $-0.3$\,pp [$-3.3$, $2.5$] ($-1.7$\,pp on the full benchmark). As multi-video accuracy is already low, the smaller effect may partly reflect a floor. Running is the hardest condition: relative to fast walking it changes single-video accuracy by $-8.1$\,pp [$-15.3$, $-0.4$], and relative to slow walking it changes multi-video accuracy by $-4.8$\,pp [$-8.4$, $-1.3$].

\noindent\textbf{Ordered route understanding is weak even within one observation.}
\label{sec:single_category_analysis}
The strongest single-video trajectory score is only $53.1\%$, achieved by Qwen3.5-27B-Thinking, which reaches $75.4\%$ on ego-relative spatial relations.
Because $91$ of the $96$ trajectory questions are single-choice, this weakness cannot be attributed solely to exact multi-select evaluation.

The same pattern recurs in multi-video evaluation (Table~\ref{tab:multi_capability}): turns and route shape form the weakest capability group for both Gemini-3.8-Flash ($64.0\%$) and Gemma-4-31B-Thinking ($36.8\%$), and Gemini-3.8-Flash obtains only $33.3\%$ on multi-step turning over whole-video evidence ($n=87$).
Such intersections are diagnostic only: several input--capability cells contain fewer than $20$ questions, and $10$ are empty.
Cross-video understanding thus depends on trajectory information that is already difficult to recover locally.

\noindent\textbf{Reasoning gains are selective across capabilities.}
\label{sec:thinking_ablation}
Across the eight matched single-video comparisons in Table~\ref{tab:single_results}, enabling reasoning improves every configuration, with a median gain of $7.8$ percentage points.
The median gain is much larger for ego-relative spatial relations ($14.0$ points) than for trajectory understanding ($3.7$ points).
Similarly, Qwen3-VL-235B improves from $42.6\%$ to $64.0\%$ on multi-select questions but only from $60.6\%$ to $62.8\%$ on single-choice questions.
Reasoning thus appears to help verify candidate relations more than reconstruct ordered motion, although these comparisons also reflect the longer outputs of reasoning runs.
We restrict this controlled comparison to the single-video split.


\begin{figure*}[t]
    \centering
    \includegraphics[width=\textwidth]{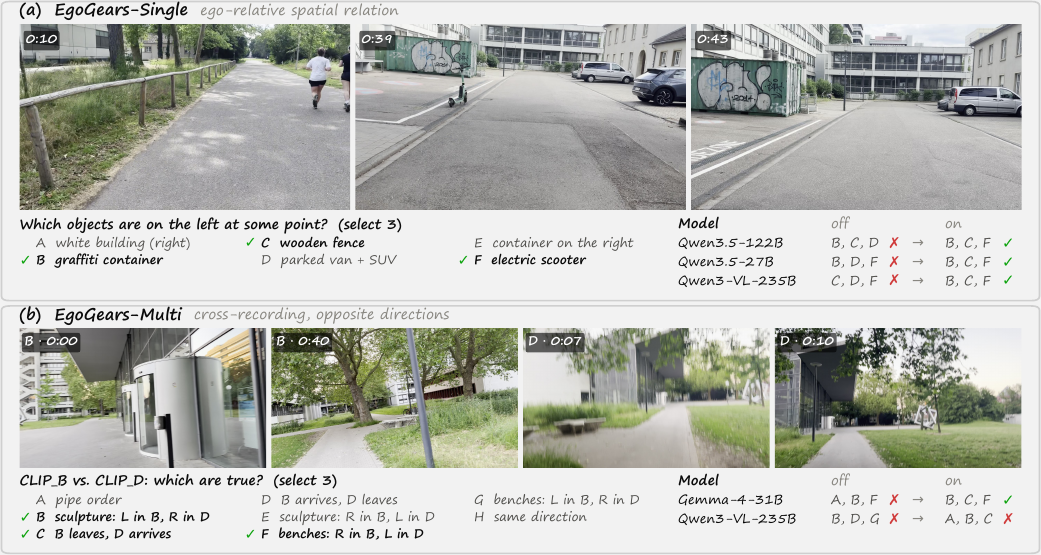}
    \caption{Qualitative cases illustrating the benefits and limitations of explicit reasoning. \textbf{(a)} On single-video ego-relative spatial QA, three configurations fail without reasoning but recover the exact three-option answer when reasoning is enabled. \textbf{(b)} Cross-recording correspondence also requires matching the same place across opposite travel directions.}
    \label{fig:qualitative_cases}
    \vspace{-3mm}
\end{figure*}

\noindent\textbf{Qualitative case study.}
Figure~\ref{fig:qualitative_cases} contrasts evidence verification with cross-observation reasoning. In panel (a), the three correct objects appear on the left at different times; every direct-answer configuration instead selects the parked vehicles on the right and omits one correct object, whereas with reasoning each model produces the exact set $\{\mathrm{B},\mathrm{C},\mathrm{F}\}$. Panel (b) shows the same place traversed in opposite directions, so a model must establish correspondence, decide whether each camera approaches or leaves the doors, and reverse the left--right relations of the sculpture and benches. Reasoning corrects Gemma-4-31B from $\{\mathrm{A},\mathrm{B},\mathrm{F}\}$ to $\{\mathrm{B},\mathrm{C},\mathrm{F}\}$, but Qwen3-VL-235B still selects $\mathrm{A}$ and omits $\mathrm{F}$. Reasoning can verify individual propositions, but cross-encounter success also requires one consistent representation of place, direction, and route phase.

\section{Conclusion}
\label{sec:conclusion}

We introduced EgoGears, a benchmark of repeated outdoor egocentric traversals
that complements local video understanding with reasoning across encounters. Current
MLLMs perceive individual observations far better than they relate them: clip--evidence
binding and ordered route understanding remain key bottlenecks, and explicit
reasoning aids relation verification more than trajectory reconstruction.

\bibliography{iclr2027_conference}
\bibliographystyle{iclr2027_conference}
\newpage
\appendix
\section*{Appendix}

\section{Societal Impact and Limitations}
\label{app:impact}

\paragraph{Societal impact.}
EgoGears evaluates whether multimodal models can relate egocentric observations of outdoor routes recorded while walking and running, under changes in speed, viewpoint, and illumination. Reliable models of this kind could support wearable route assistants, assistive navigation for people with visual impairments, place recognition for robots that learn from human route demonstrations, and the analysis of outdoor mobility. By separating local understanding from cross-video correspondence, the benchmark shows where current models fail, for example on ordered turns and on binding evidence to the correct clip, before such systems are deployed.

The same capabilities raise risks. Recordings made in public spaces may contain bystanders, vehicles, and private property. Released clips are downscaled to 480p without audio, and anonymized identifiers conceal participants, routes, and recording conditions. 
Before release, faces and license plates will be blurred to reduce the identifiability of participants and incidental bystanders. Tasks that recognize the same place or route across recordings could, in principle, be used to infer where a person moves from videos they share. We therefore intend the benchmark for research on video understanding and not for surveillance or re-identification. In addition, higher accuracy on EgoGears does not show that a model is safe for real-time use: a wrong answer about a turn, an obstacle, or a crossing could endanger a pedestrian or runner. Deployment would require human oversight, calibrated uncertainty, and validation in the target environment.

\paragraph{Limitations.}
First, all $126$ recordings come from a small number of participants and locations in one country, so they may not represent other regions, climates, terrains, body types, or recording devices. Second, the speed--lighting design is fully crossed on only $9$ of the $39$ routes, and night recordings cover $16$ routes; the condition analysis in Figure~\ref{fig:condition-bar} therefore rests on a small controlled subset and should be read as descriptive. Third, candidate questions were generated and partly verified with Gemini models, and Gemini also participated in adjudicating part of the multi-video set. Human review and the checks in Appendix~\ref{sec:reproducibility} reduce but do not eliminate residual label noise and possible model-specific bias. Fourth, results depend on the evaluation protocol: long reasoning outputs can be truncated, so some failures reflect incomplete outputs rather than a lack of understanding. Finally, the benchmark uses closed-set multiple-choice questions and offline clips; it does not evaluate open-ended answers or streaming, real-time interaction. Future work should extend the controlled design to more routes, participants, and regions, add conditions such as rain, snow, and crowded scenes, and evaluate models in open-ended and online settings.

\section{Implementation Details and Model Configurations}
\label{app:implementation}
\subsection{Models and Configurations}

We evaluate configurations from the Qwen2.5-VL, Qwen3-VL, Qwen3.5, InternVL3.5, Gemma, GLM, Kimi-VL, together with three Gemini API models on the multi-video split. A configuration denotes a checkpoint and an inference mode, so reasoning-enabled and direct-answer modes of the same checkpoint are evaluated separately. The single-video leaderboard reports 29 configurations (Table~\ref{tab:single_results}). The multi-video leaderboard reports 31 configurations (Table~\ref{tab:multi_results}): 3 Gemini API models, 18 open-source base runs, and 10 open-source explicit-reasoning runs. For the four checkpoints released only as Thinking models (Qwen3-VL-235B-A22B-Thinking, Qwen3-VL-30B-A3B-Thinking, GLM-4.1V-9B-Thinking, and Kimi-VL-A3B-Thinking-2506), we report only the explicit-reasoning run of each checkpoint.

\subsection{Visual Input and Prompting}

Models receive the evaluation clips of each question as images without audio. Non-contiguous clips and clips from different recordings remain separate rather than being concatenated into an artificial video. Each clip is introduced by an anonymized label.

Option order is deterministically shuffled using the question identifier and mapped back before scoring. All models receive the same question, shuffled options, and required selection count. The prompt instructs the model to use visual evidence, select exactly the requested number of options, and return only a JSON object of the form \texttt{\{"answer": [...]\}}. Decoding is deterministic with temperature zero. The multi-video prompt differs only in introducing each independent clip with its label; the single-video prompt states that the frames are in chronological order.

\subsection{Inference and Answer Parsing}

The default model context length is 65,536 tokens. InternVL3.5 uses its 40,960-token limit. All InternVL3.5 models except 20B-A4B use one visual tile per image, because default dynamic tiling would exceed this limit.

Single-video open models are served through an OpenAI-compatible interface using vLLM 0.30.0, Transformers 5.17.0, and PyTorch 2.13.0 with CUDA 13.0. We explicitly set the reasoning switch for models that support both modes and use checkpoint defaults for fixed Instruct or Thinking models. Model-specific reasoning parsers are used where required. In particular, Qwen models use the Qwen3 parser, Gemma-4 reasoning runs use the Gemma-4 parser, and Kimi's custom reasoning delimiters are separated by the evaluation runner.

The scorer removes the reasoning channel and accepts only the last post-reasoning JSON object containing an \texttt{answer} key. It provides no free-text fallback, and letters outside the displayed option range are invalid. Truncated, unparseable, missing, or incomplete predictions remain in the fixed benchmark denominator and are counted as incorrect. Each single-video run records dataset and prompt hashes so that incompatible configurations cannot be resumed accidentally.

Multi-video runs use the same prompts, visual input, and option shuffling. The multi-video results indicate overall difficulty and differences between model groups, but differences of a few points among open-source configurations should not be read as capability rankings.

\subsection{Metrics}
\label{app:metrics}

The primary metric is exact-set accuracy: a prediction is correct only if its selected option set exactly equals the reference set. We additionally report exact accuracy separately for single-choice and multi-select questions (Table~\ref{tab:single_results}) and a partial-credit score, the mean per-question Jaccard similarity $|P\cap G|/|P\cup G|$ between the predicted set $P$ and the reference set $G$, which penalizes both missing and extra options. To compare subsets with different answer formats, we use chance-normalized accuracy
\begin{equation}
    \kappa = \frac{a-c}{1-c},
\end{equation}
where $a$ is observed exact accuracy and $c$ is the exact-set accuracy of uniform random selection given each question's option and selection counts, averaged over the subset. On the 531 questions spanning independent recordings, the random baseline is 4.47\%.

Table~\ref{tab:partial_credit} reports Jaccard and $\kappa_{531}$ for all configurations. Partial credit raises scores by a median of 9.2 points on single-video and 18.3 points on multi-video questions, but it barely changes the ranking of configurations (Spearman $\rho \geq 0.98$ on both splits). Conclusions drawn from exact-set accuracy therefore do not depend on the strictness of the metric.

\begin{table}[!htb]
\centering
\caption{\textbf{Exact-set versus partial-credit accuracy} (\%). Jaccard is the mean per-question $|P\cap G|/|P\cup G|$ between the predicted set $P$ and the reference set $G$; failed or missing answers score $0$. $\kappa_{531}$ is chance-normalized exact accuracy on the $531$ questions spanning independent recordings (random baseline $4.47\%$). Configurations and settings match Tables~\ref{tab:single_results} and~\ref{tab:multi_results}; -- marks a configuration not evaluated on that split.}
\label{tab:partial_credit}
\begingroup
\fontsize{7.2}{8.2}\selectfont
\setlength{\tabcolsep}{4pt}
\renewcommand{\arraystretch}{1.12}
\begin{tabular}{@{}lcrrrrr@{}}
\toprule
& & \multicolumn{2}{c}{Single-video ($N=567$)} & \multicolumn{3}{c}{Multi-video ($N=1{,}487$)} \\
\cmidrule(lr){3-4}\cmidrule(l){5-7}
Model & Mode & Exact & Jaccard & Exact & Jaccard & $\kappa_{531}$ \\
\midrule
\rowcolor{egGemini}\multicolumn{7}{@{}l}{\textit{Gemini (API)}}\\
\rowcolor{egGemini}3.8-Flash & API & -- & -- & \textbf{78.3} & \textbf{83.1} & \textbf{82.7} \\
\rowcolor{egGemini}3.1-Pro & API & -- & -- & 66.5 & 76.5 & 68.1 \\
\rowcolor{egGemini}2.5-Pro & API & -- & -- & 56.4 & 68.6 & 55.4 \\
\addlinespace[2pt]
\rowcolor{egQwen35}\multicolumn{7}{@{}l}{\textit{Qwen3.5}}\\
\rowcolor{egQwen35}122B-A10B & T & \textbf{68.6} & 74.1 & 37.8 & 53.9 & 26.5 \\
\rowcolor{egQwen35} & D & 59.1 & 66.4 & 16.3 & 38.4 & 2.4 \\
\rowcolor{egQwen35}35B-A3B & T & 66.5 & 70.9 & 37.9 & 53.6 & 28.0 \\
\rowcolor{egQwen35} & D & 60.8 & 68.1 & 15.1 & 37.3 & 0.8 \\
\rowcolor{egQwen35}27B & T & \textbf{68.6} & 73.1 & 40.8 & 56.3 & 34.4 \\
\rowcolor{egQwen35} & D & 62.6 & 69.7 & 16.0 & 38.5 & 5.0 \\
\rowcolor{egQwen35}9B & T & 58.6 & 63.0 & 21.3 & 38.4 & 11.5 \\
\rowcolor{egQwen35} & D & 48.5 & 57.7 & 10.9 & 30.4 & 0.4 \\
\addlinespace[2pt]
\rowcolor{egQwen3VL}\multicolumn{7}{@{}l}{\textit{Qwen3-VL}}\\
\rowcolor{egQwen3VL}235B-A22B & T & 63.3 & 69.8 & 46.0 & 61.5 & 36.1 \\
\rowcolor{egQwen3VL} & D & 52.9 & 62.0 & 28.6 & 47.9 & 20.4 \\
\rowcolor{egQwen3VL}30B-A3B & T & 56.1 & 64.1 & 26.2 & 44.8 & 15.2 \\
\rowcolor{egQwen3VL}8B & T & 52.6 & 59.3 & -- & -- & -- \\
\rowcolor{egQwen3VL} & D & 40.7 & 52.3 & 16.9 & 35.6 & 7.3 \\
\addlinespace[2pt]
\rowcolor{egQwen25VL}\multicolumn{7}{@{}l}{\textit{Qwen2.5-VL}}\\
\rowcolor{egQwen25VL}72B & D & 51.7 & 62.0 & 26.4 & 45.7 & 19.8 \\
\rowcolor{egQwen25VL}7B & D & 22.4 & 33.9 & 9.8 & 26.7 & 1.4 \\
\addlinespace[2pt]
\rowcolor{egGemma}\multicolumn{7}{@{}l}{\textit{Gemma}}\\
\rowcolor{egGemma}4-31B & T & 68.4 & \textbf{74.4} & 56.2 & 69.5 & 53.5 \\
\rowcolor{egGemma} & D & 62.3 & 69.6 & 33.2 & 52.5 & 25.5 \\
\rowcolor{egGemma}4-26B-A4B & D & 49.0 & 58.8 & 22.0 & 43.2 & 13.3 \\
\rowcolor{egGemma}3n-E4B & D & 19.8 & 31.0 & 9.0 & 25.8 & 3.0 \\
\addlinespace[2pt]
\rowcolor{egIntern}\multicolumn{7}{@{}l}{\textit{InternVL3.5}}\\
\rowcolor{egIntern}241B-A28B & D & 45.7 & 57.7 & 24.2 & 43.9 & 14.8 \\
\rowcolor{egIntern}38B & D & 27.5 & 40.5 & 18.1 & 38.5 & 10.7 \\
\rowcolor{egIntern}30B-A3B & D & 27.3 & 40.0 & 13.4 & 31.7 & 5.8 \\
\rowcolor{egIntern}20B-A4B & D & 39.9 & 50.4 & 13.9 & 31.9 & 3.6 \\
\rowcolor{egIntern}14B & D & 36.3 & 48.9 & 16.5 & 36.9 & 7.2 \\
\rowcolor{egIntern}8B & D & 27.0 & 39.9 & 10.4 & 28.5 & 2.8 \\
\addlinespace[2pt]
\rowcolor{egGLM}\multicolumn{7}{@{}l}{\textit{GLM}}\\
\rowcolor{egGLM}4.6V & T & 51.1 & 59.0 & 39.9 & 56.9 & 38.5 \\
\rowcolor{egGLM} & D & 45.9 & 55.7 & 36.9 & 54.9 & 34.6 \\
\rowcolor{egGLM}4.1V-9B & T & 41.3 & 51.3 & 13.4 & 35.4 & 12.1 \\
\addlinespace[2pt]
\rowcolor{egKimi}\multicolumn{7}{@{}l}{\textit{Kimi}}\\
\rowcolor{egKimi}VL-A3B-2506 & T & 45.7 & 57.4 & 19.5 & 41.3 & 8.3 \\
\bottomrule
\end{tabular}
\endgroup
\end{table}

\subsection{Single- versus Multi-video Comparison}
\label{app:split_gap}

Twenty configurations are evaluated in comparable modes on both splits. Both splits are drawn from the same pool of $126$ recordings, but they differ in answer format: the share of multi-select questions and the option and selection counts vary. Table~\ref{tab:split_gap} therefore compares the splits with the answer format held fixed, by restricting both to single-choice or to multi-select questions, and with partial credit instead of exact-set scoring. Every configuration scores lower on multi-video questions under all four comparisons. Format and scoring account for part of the overall gap, which falls from a mean of $22.5$ points to $14.6$ points among single-choice questions and to $13.3$ points under partial credit, but the gap remains positive for every configuration.

\begin{table}[!htb]
\centering
\caption{\textbf{Single- versus multi-video accuracy} (\%) for the $20$ configurations evaluated in comparable modes on both splits. Both splits are drawn from the same recordings. SC and MS restrict each split to its single-choice or multi-select questions; Jaccard is the partial-credit score of Table~\ref{tab:partial_credit}. $\Delta$ is the single-video minus multi-video score.}
\label{tab:split_gap}
\begingroup
\fontsize{7.2}{8.2}\selectfont
\setlength{\tabcolsep}{3.2pt}
\renewcommand{\arraystretch}{1.1}
\begin{tabular}{@{}lcrrrrrrrrrrrr@{}}
\toprule
& & \multicolumn{3}{c}{All (exact)} & \multicolumn{3}{c}{SC (exact)} & \multicolumn{3}{c}{MS (exact)} & \multicolumn{3}{c}{All (Jaccard)} \\
\cmidrule(lr){3-5}\cmidrule(lr){6-8}\cmidrule(lr){9-11}\cmidrule(l){12-14}
Model & Mode & S & M & $\Delta$ & S & M & $\Delta$ & S & M & $\Delta$ & S & M & $\Delta$ \\
\midrule
Qwen3.5-27B & T & 68.6 & 40.8 & 27.8 & 69.5 & 45.2 & 24.3 & 67.4 & 38.7 & 28.7 & 73.1 & 56.3 & 16.8 \\
Qwen3.5-122B-A10B & T & 68.6 & 37.8 & 30.8 & 68.9 & 42.7 & 26.2 & 68.2 & 35.5 & 32.7 & 74.1 & 53.9 & 20.2 \\
Qwen3.5-35B-A3B & T & 66.5 & 37.9 & 28.6 & 67.7 & 46.1 & 21.6 & 64.9 & 34.1 & 30.8 & 70.9 & 53.6 & 17.3 \\
Qwen3.5-9B & T & 58.6 & 21.3 & 37.3 & 59.7 & 34.6 & 25.1 & 57.0 & 15.2 & 41.8 & 63.0 & 38.4 & 24.6 \\
Gemma-4-31B & T & 68.4 & 56.2 & 12.2 & 69.5 & 62.8 & 6.7 & 66.9 & 53.1 & 13.8 & 74.4 & 69.5 & 4.9 \\
GLM-4.6V & T & 51.1 & 39.9 & 11.2 & 51.4 & 46.9 & 4.5 & 50.8 & 36.6 & 14.2 & 59.0 & 56.9 & 2.1 \\
Qwen3-VL-235B-A22B & T & 63.3 & 46.0 & 17.3 & 62.8 & 56.3 & 6.5 & 64.0 & 41.2 & 22.8 & 69.8 & 61.5 & 8.3 \\
Qwen3-VL-30B-A3B & T & 56.1 & 26.2 & 29.9 & 58.2 & 48.2 & 10.0 & 53.3 & 15.9 & 37.4 & 64.1 & 44.8 & 19.3 \\
GLM-4.1V-9B & T & 41.3 & 13.4 & 27.9 & 47.1 & 24.8 & 22.3 & 33.5 & 8.2 & 25.3 & 51.3 & 35.4 & 15.9 \\
Kimi-VL-A3B-2506 & T & 45.7 & 19.5 & 26.2 & 55.4 & 41.4 & 14.0 & 32.6 & 9.4 & 23.2 & 57.4 & 41.3 & 16.1 \\
Qwen3-VL-235B-A22B & D & 52.9 & 28.6 & 24.3 & 60.6 & 49.0 & 11.6 & 42.6 & 19.1 & 23.5 & 62.0 & 47.9 & 14.1 \\
Qwen3-VL-8B & D & 40.7 & 16.9 & 23.8 & 54.5 & 37.2 & 17.3 & 22.3 & 7.5 & 14.8 & 52.3 & 35.6 & 16.7 \\
Qwen2.5-VL-72B & D & 51.7 & 26.4 & 25.3 & 60.3 & 43.1 & 17.2 & 40.1 & 18.6 & 21.5 & 62.0 & 45.7 & 16.3 \\
Qwen2.5-VL-7B & D & 22.4 & 9.8 & 12.6 & 35.1 & 23.8 & 11.3 & 5.4 & 3.3 & 2.1 & 33.9 & 26.7 & 7.2 \\
Gemma-4-26B-A4B & D & 49.0 & 22.0 & 27.0 & 55.1 & 39.3 & 15.8 & 40.9 & 14.0 & 26.9 & 58.8 & 43.2 & 15.6 \\
Gemma-3n-E4B & D & 19.8 & 9.0 & 10.8 & 28.9 & 20.2 & 8.7 & 7.4 & 3.8 & 3.6 & 31.0 & 25.8 & 5.2 \\
InternVL3.5-20B-A4B & D & 39.9 & 13.9 & 26.0 & 55.4 & 32.3 & 23.1 & 19.0 & 5.4 & 13.6 & 50.4 & 31.9 & 18.5 \\
InternVL3.5-241B-A28B & D & 45.7 & 24.2 & 21.5 & 58.8 & 45.6 & 13.2 & 28.1 & 14.3 & 13.8 & 57.7 & 43.9 & 13.8 \\
InternVL3.5-38B & D & 27.5 & 18.1 & 9.4 & 35.4 & 35.0 & 0.4 & 16.9 & 10.2 & 6.7 & 40.5 & 38.5 & 2.0 \\
InternVL3.5-14B & D & 36.3 & 16.5 & 19.8 & 51.1 & 38.9 & 12.2 & 16.5 & 6.1 & 10.4 & 48.9 & 36.9 & 12.0 \\
\midrule
\multicolumn{2}{@{}l}{Mean} & 48.7 & 26.2 & 22.5 & 55.3 & 40.7 & 14.6 & 39.9 & 19.5 & 20.4 & 57.7 & 44.4 & 13.3 \\
\multicolumn{2}{@{}l}{Min.\ $\Delta$} & \multicolumn{3}{r}{9.4} & \multicolumn{3}{r}{0.4} & \multicolumn{3}{r}{2.1} & \multicolumn{3}{r}{2.0} \\
\bottomrule
\end{tabular}
\endgroup
\end{table}

\subsection{Evaluation Prompts}

All evaluations use the two fixed templates below, unchanged between direct-answer and reasoning modes. They differ only in how the visual input is introduced. The wording is reproduced exactly; only line breaks are adjusted for layout. The visual input follows the text as images. In the multi-video setting, each clip's frames are preceded by a text part \texttt{--- CLIP\_X ---} that carries the clip's anonymized label. The placeholders are filled as listed below the templates.
\begin{center}
\begin{minipage}[t]{0.485\linewidth}
\begin{promptbox}[equal height group=prompts]{Single-video prompt}
Answer this multiple-choice question about egocentric walking/running footage.
The frames below are sampled in chronological order from the single video clip (at most 60 seconds long) associated with this question.

Watch the video frames before deciding. Base your answer on visual evidence from this video. Do not answer from the wording of the options alone.

QUESTION: \ph{question}

OPTIONS:\\\ph{options}

Select exactly \ph{n} option\ph{plural}.\\
Return ONLY JSON: \{\dq answer\dq: [\ph{example}]\}
\end{promptbox}
\end{minipage}\hfill
\begin{minipage}[t]{0.485\linewidth}
\begin{promptbox}[equal height group=prompts]{Multi-video prompt}
Answer this multiple-choice question about egocentric walking/running footage.
The frames below are sampled in order from the clip(s) the question refers to; each clip is introduced by its label.

Watch before deciding. Do not answer from the wording of the options alone.

QUESTION: \ph{question}

OPTIONS:\\\ph{options}

Select exactly \ph{n} option\ph{plural}.\\
Return ONLY JSON: \{\dq answer\dq: [\ph{example}]\}
\end{promptbox}
\end{minipage}
\end{center}

\begin{center}
\small
\begin{tabularx}{0.94\linewidth}{@{}lX@{}}
\toprule
Placeholder & Filled with \\
\midrule
\ph{question} & The question stem. \\
\ph{options} & The options after per-question shuffling, one per line as \texttt{A.~\textit{text}}. \\
\ph{n}, \ph{plural} & The required number of selections; \texttt{s} when $n>1$, otherwise empty. \\
\ph{example} & The first $n$ displayed option letters, e.g., \texttt{\dq A\dq, \dq B\dq, \dq C\dq}. Because options are shuffled, copying the example does not favor the reference answer. \\
\bottomrule
\end{tabularx}
\end{center}

\subsection{Evaluation Taxonomies}

Multi-video questions have two complementary classification axes. The \emph{unit} axis describes how evidence is organized: whole-video, multi-caption multi-clip, anonymous multi-clip, cross-video, and cross-recording. Whole-video and multi-caption questions draw clips from one recording, whereas anonymous, cross-video, and cross-recording questions require evidence from independent recordings. The latter three units contain 531 questions in total.

The \emph{capability} axis merges 29 fine-grained question types into seven groups: landmark recognition and ordering, event and temporal reasoning, turns and route shape, global route and phase understanding, spatial relations, cross-recording route matching, and environmental change. Each group contains at least 113 questions. We report the two axes separately because they are not independent: 10 of the 35 unit--capability intersections are empty, and several non-empty intersections contain fewer than 20 questions. Table~\ref{tab:multi_results} reports accuracy along the unit axis, and Table~\ref{tab:multi_capability} along the capability axis.

\begin{table}[!htb]
\centering
\caption{\textbf{Multi-video QA by capability} ($N=1{,}487$). Exact-set accuracy (\%) on the seven capability groups, with question counts in the header. MST is the multi-step-turns question type over whole-video evidence ($n=87$), a subset of Turn. Configurations and modes match Table~\ref{tab:multi_results}.}
\label{tab:multi_capability}
\begingroup
\fontsize{7.2}{8.2}\selectfont
\setlength{\tabcolsep}{4pt}
\renewcommand{\arraystretch}{1.12}
\begin{tabular}{@{}lcrrrrrrrr@{}}
\toprule
Model & Mode & \makecell{Land.\\(337)} & \makecell{Event\\(287)} & \makecell{Turn\\(239)} & \makecell{Route\\(239)} & \makecell{Spatial\\(154)} & \makecell{Match\\(118)} & \makecell{Env.\\(113)} & \makecell{MST\\(87)} \\
\midrule
\rowcolor{white}Random & -- & 7.3 & 4.9 & 6.8 & 10.4 & 2.4 & 6.6 & 6.6 & 15.5 \\
\midrule
\rowcolor{egGemini}\multicolumn{10}{@{}l}{\textit{Gemini (API)}}\\
\rowcolor{egGemini}3.8-Flash & API & \textbf{79.8} & \textbf{77.7} & \textbf{64.0} & \textbf{82.0} & \textbf{85.7} & \textbf{85.6} & \textbf{79.6} & \textbf{33.3} \\
\rowcolor{egGemini}3.1-Pro & API & 67.7 & 65.9 & 50.6 & 77.4 & 66.2 & 73.7 & 68.1 & 29.9 \\
\rowcolor{egGemini}2.5-Pro & API & 53.7 & 55.1 & 39.7 & 71.1 & 60.4 & 61.0 & 61.9 & 16.1 \\
\addlinespace[2pt]
\rowcolor{egQwen35}\multicolumn{10}{@{}l}{\textit{Qwen3.5}}\\
\rowcolor{egQwen35}122B-A10B & T & 36.8 & 39.0 & 29.7 & 51.5 & 34.4 & 23.7 & 45.1 & 18.4 \\
\rowcolor{egQwen35} & D & 17.2 & 20.2 & 5.9 & 31.8 & 6.5 & 6.8 & 16.8 & 4.6 \\
\rowcolor{egQwen35}35B-A3B & T & 37.4 & 41.1 & 20.9 & 54.0 & 39.6 & 24.6 & 44.2 & 10.3 \\
\rowcolor{egQwen35} & D & 16.0 & 19.5 & 5.0 & 28.5 & 5.8 & 3.4 & 19.5 & 6.9 \\
\rowcolor{egQwen35}27B & T & 45.1 & 38.7 & 27.2 & 55.6 & 43.5 & 31.4 & 36.3 & 18.4 \\
\rowcolor{egQwen35} & D & 21.1 & 14.6 & 3.8 & 29.7 & 9.1 & 6.8 & 20.4 & 3.4 \\
\rowcolor{egQwen35}9B & T & 21.1 & 20.9 & 10.9 & 36.4 & 14.3 & 18.6 & 25.7 & 12.6 \\
\rowcolor{egQwen35} & D & 13.4 & 8.0 & 2.9 & 23.4 & 3.9 & 7.6 & 14.2 & 3.4 \\
\addlinespace[2pt]
\rowcolor{egQwen3VL}\multicolumn{10}{@{}l}{\textit{Qwen3-VL}}\\
\rowcolor{egQwen3VL}235B-A22B & T & 49.6 & 51.9 & 28.9 & 59.4 & 39.6 & 38.1 & 45.1 & 24.1 \\
\rowcolor{egQwen3VL} & D & 27.0 & 26.5 & 21.3 & 46.4 & 14.9 & 34.7 & 28.3 & 28.7 \\
\rowcolor{egQwen3VL}30B-A3B & T & 25.8 & 25.4 & 13.8 & 42.3 & 22.7 & 21.2 & 31.0 & 14.9 \\
\rowcolor{egQwen3VL}8B & D & 18.4 & 13.9 & 12.1 & 28.0 & 5.8 & 17.8 & 20.4 & 21.8 \\
\addlinespace[2pt]
\rowcolor{egQwen25VL}\multicolumn{10}{@{}l}{\textit{Qwen2.5-VL}}\\
\rowcolor{egQwen25VL}72B & D & 28.5 & 23.0 & 15.9 & 41.4 & 13.6 & 32.2 & 30.1 & 21.8 \\
\rowcolor{egQwen25VL}7B & D & 8.6 & 7.3 & 5.0 & 23.0 & 5.8 & 9.3 & 8.0 & 11.5 \\
\addlinespace[2pt]
\rowcolor{egGemma}\multicolumn{10}{@{}l}{\textit{Gemma}}\\
\rowcolor{egGemma}4-31B & T & 56.4 & 55.1 & 36.8 & 71.1 & 57.1 & 61.9 & 60.2 & 27.6 \\
\rowcolor{egGemma} & D & 33.8 & 30.7 & 22.6 & 50.6 & 26.6 & 27.1 & 38.1 & 24.1 \\
\rowcolor{egGemma}4-26B-A4B & D & 24.6 & 21.6 & 13.4 & 32.6 & 10.4 & 21.2 & 27.4 & 17.2 \\
\rowcolor{egGemma}3n-E4B & D & 9.5 & 4.2 & 10.5 & 17.2 & 1.9 & 8.5 & 9.7 & 16.1 \\
\addlinespace[2pt]
\rowcolor{egIntern}\multicolumn{10}{@{}l}{\textit{InternVL3.5}}\\
\rowcolor{egIntern}241B-A28B & D & 29.7 & 20.6 & 15.9 & 37.2 & 10.4 & 26.3 & 23.9 & 23.0 \\
\rowcolor{egIntern}38B & D & 19.0 & 19.9 & 13.4 & 24.7 & 9.1 & 20.3 & 16.8 & 18.4 \\
\rowcolor{egIntern}30B-A3B & D & 14.5 & 11.8 & 7.9 & 23.0 & 6.5 & 16.1 & 11.5 & 14.9 \\
\rowcolor{egIntern}20B-A4B & D & 13.4 & 10.1 & 6.7 & 30.5 & 9.7 & 11.0 & 14.2 & 13.8 \\
\rowcolor{egIntern}14B & D & 16.0 & 14.3 & 10.9 & 30.5 & 7.1 & 15.3 & 19.5 & 17.2 \\
\rowcolor{egIntern}8B & D & 9.8 & 8.0 & 9.6 & 18.4 & 7.8 & 10.2 & 7.1 & 13.8 \\
\addlinespace[2pt]
\rowcolor{egGLM}\multicolumn{10}{@{}l}{\textit{GLM}}\\
\rowcolor{egGLM}4.6V & T & 37.7 & 35.5 & 31.8 & 51.5 & 39.0 & 51.7 & 38.9 & 23.0 \\
\rowcolor{egGLM} & D & 39.5 & 31.4 & 27.2 & 48.5 & 36.4 & 40.7 & 35.4 & 23.0 \\
\rowcolor{egGLM}4.1V-9B & T & 11.6 & 9.4 & 9.6 & 18.8 & 13.6 & 25.4 & 13.3 & 13.8 \\
\addlinespace[2pt]
\rowcolor{egKimi}\multicolumn{10}{@{}l}{\textit{Kimi}}\\
\rowcolor{egKimi}VL-A3B-2506 & T & 19.3 & 16.0 & 12.1 & 35.1 & 11.7 & 16.9 & 24.8 & 19.5 \\
\bottomrule
\end{tabular}
\endgroup
\par\vspace{3pt}
\begin{minipage}{\linewidth}\fontsize{7}{8.2}\selectfont Land.\ = landmark recognition and ordering; Event = event and temporal reasoning; Turn = turns and route shape; Route = global route and phase understanding; Spatial = spatial relations; Match = cross-recording route matching; Env.\ = environmental change. Random is the exact-set accuracy of uniform random selection. Bold marks the best score in each column.\end{minipage}
\end{table}

Single-video questions use nine semantic categories: object and attribute QA, ego-relative spatial relation, trajectory-grounded QA, ego-motion recognition, scene and place QA, text and OCR, temporal grounding, actor action and motion, and event sequencing. The main table emphasizes the first three categories because each contains at least 30 questions. The remaining categories are retained as high-variance diagnostics rather than used for fine-grained model ranking.

\section{Collection and Quality-Control Details}
\label{app:collection_details}

\subsection{Acquisition and Evidence Preparation}
\label{app:video_acquisition}

Videos were captured using smartphones at 1080p and $30$ fps. Participants mounted the phone at chest level, with its plane perpendicular to the ground and the camera facing forward. Collection instructions specified continuous sessions of at least $10$ minutes. Route selection prioritized recognizable landmarks, limited pedestrian presence, and varied geometry, including straight segments, left and right turns, and U-turns.

The acquisition protocol targeted repeated traversals across movement and lighting conditions. Realized coverage is partial: $54$ recordings are explicitly identified as out-and-back traversals, and $9$ routes cover all six movement--lighting combinations. 

Evaluation clips are encoded as 480p H.264 video with audio removed. Evidence preparation follows each question's temporal scope: continuous excerpts are extracted directly, while questions about complete recordings may use multiple evidence windows from the same source. Non-contiguous windows and clips from independent recordings remain separate files.

Anonymized clip labels and paths conceal original filenames, participant identities, route identifiers, movement and lighting metadata, and recording identities. Internal mappings retain the association between each question, its source recordings, and its evidence windows.

\subsection{Model-Assisted Candidate Generation}
\label{app:candidate_generation}

We use gemini-3.1-pro-preview for video observation and visual verification. For EgoGears-Multi, gemini-3.5-flash generates question stems, options, proposed answers, question types, and evidence ranges from the structured observations. For EgoGears-Single, gemini-3.1-pro-preview generates both captions and candidate questions. Text-only screening uses gemini-3.5-flash throughout.

For EgoGears-Multi, observation scope follows the question: complete recordings support global route questions, continuous excerpts support local temporal questions, and joint observations of independent recordings support cross-video comparisons. Segment-based generation uses clips at multiple temporal scales, including $60$ and $180$ seconds. Structured observations describe route phases, turns, landmarks, repeated places, dynamic events, environmental changes, and spatial relations. This process produces approximately $11{,}802$ candidate questions.

For EgoGears-Single, recordings are divided into $60$-second clips, retaining shorter tail clips. Each clip receives a visually grounded caption with structured fields for the environment, objects and attributes, spatial relations, camera trajectories, and events. Candidate questions are generated from these descriptions.

\subsection{Automated Screening for EgoGears-Multi}
\label{app:multi_screening}

\paragraph{Whole-recording and continuous-excerpt questions.}
These candidates pass a five-stage procedure:
\begin{enumerate}
    \item \textbf{Deterministic option shuffling.}
    Option positions are randomized using a stable procedure.

    \item \textbf{Structural validation.}
    Checks enforce consistency among the stem, options, answer labels, question type, and selection count, and reject duplicate options.

    \item \textbf{Leakage screening.}
    Questions are screened for identifying metadata, internal annotation fields, caption-derived cues, and timestamps that reveal the answer.

    \item \textbf{Text-only screening.}
    The text-only model answers each question three times without video. A question is flagged if at least two responses match the proposed answer, triggering rejection or rewriting. This is a heuristic screen for language or option-structure shortcuts.

    \item \textbf{Video-grounded verification.}
    The video model checks the required footage against every option, identifying unsupported claims, directional errors, and inconsistencies between the question and its evidence.
\end{enumerate}

\paragraph{Multi-clip and cross-video questions.}
These candidates pass an eight-stage procedure:
\begin{enumerate}
    \item \textbf{Comparison requirement.}
    The question must belong to an allowed task type, contain sufficient clips, and require reasoning across multiple evidence windows.

    \item \textbf{Option structure.}
    Checks examine option and answer counts, option length, and information density for superficial answer cues.

    \item \textbf{Leakage screening.}
    Identifying metadata, internal fields, and description-dependent wording such as \texttt{according to the description} are screened out.

    \item \textbf{Deterministic shuffling.}
    Options and clip display order are randomized so that answer positions and alphabetic clip labels do not encode the intended relation.

    \item \textbf{Cross-batch deduplication.}
    Checks identify exact and near-duplicate stems, repeated evidence patterns, and duplicate question signatures.

    \item \textbf{Text-only screening and rewriting.}
    Candidates undergo the three-attempt text-only test. Flagged questions may be rewritten and retested; persistently guessable questions are removed.

    \item \textbf{Evidence availability.}
    Required recordings and time ranges must be valid, extractable, and sufficient for review.

    \item \textbf{Answer-blind visual verification.}
    The video model receives the question and evidence without captions or proposed answers and labels each option as \texttt{supported}, \texttt{ruled\_out}, or \texttt{undecidable}. Acceptance requires sufficient evidence and agreement between the supported-option set and the proposed answer.
\end{enumerate}

Shuffling precedes text-only screening, and inexpensive structural and textual checks precede video verification. Automated screening produces a human-review package of $1{,}526$ questions.

\subsection{Verification and Sampling for EgoGears-Single}
\label{app:single_screening}

\paragraph{Pre-verification filtering.}
Candidate questions undergo the same three-attempt text-only screening, with rewriting and retesting when necessary. The resulting pool contains $6{,}770$ questions, averaging $4.8$ questions per clip. Removing $318$ questions associated with unsuitable clips, including truncated footage or insufficient visual evidence, leaves $6{,}452$ candidates.

\paragraph{Question taxonomy.}
Each question receives one primary label from nine categories:
\begin{enumerate}
    \item \textbf{Text and OCR:} reading visible text, numbers, or signs.
    \item \textbf{Temporal Grounding:} identifying when an event or object appears.
    \item \textbf{Event Sequencing:} ordering events, landmarks, or actions.
    \item \textbf{Scene and Place QA:} recognizing environmental and place properties.
    \item \textbf{Object and Attribute QA:} identifying objects and their static attributes.
    \item \textbf{Ego-relative Spatial Relation:} locating entities relative to the camera or route.
    \item \textbf{Trajectory-grounded QA:} understanding overall paths, travel directions, and turn sequences.
    \item \textbf{Ego-motion Recognition:} recognizing local camera or wearer motion.
    \item \textbf{Actor Action and Motion:} recognizing the actions of external entities.
\end{enumerate}

The classifier first uses question form to identify OCR, temporal-grounding, and sequencing questions. Other categories use alignment between the proposed correct options and caption fields. Motion questions are distinguished by whether they concern the overall route, local camera motion, or external actors. Spatial-versus-attribute ambiguity is further examined using the objects mentioned in distractors and captions; unresolved cases default to Object and Attribute QA.

\paragraph{Exhaustive visual verification.}
The verifier receives the clip, question, options, and proposed answer; captions are withheld. It assigns option-level support labels and records evidence time ranges. A question is labeled \texttt{answer\_correct} when all proposed correct options are supported and distractors are ruled out or undecidable. It is labeled \texttt{answer\_wrong} when a proposed correct option is contradicted or a distractor is supported, and \texttt{insufficient} when the evidence is inadequate.

Verification of all $6{,}452$ candidates yields $5{,}081$ \texttt{answer\_correct}, $1{,}364$ \texttt{answer\_wrong}, and $7$ \texttt{insufficient} outcomes.

\paragraph{Stratified selection.}
Only \texttt{answer\_correct} candidates enter quota-based sampling, targeting approximately $700$ questions across the nine categories. Larger categories are capped, while categories below quota contribute all passing questions. Sampling uses a fixed random seed and a strict limit of $15$ questions per source recording. Clip reuse is avoided where possible; this preference may be relaxed to fill quotas, while the recording-level cap remains fixed. The resulting $698$ questions proceed to human review.

\subsection{Human Review, Adjudication, and Release}
\label{app:human_review}

\paragraph{Review protocol.}
Each subset's review package is divided into ten batches assigned to ten annotators. For EgoGears-Multi, proposed answers are withheld. Reviewers inspect every associated clip, construct separate clip timelines, and label each option as \texttt{supported}, \texttt{ruled\_out}, or \texttt{undecidable}. They provide an answer, a clarity judgment, and notes on problematic options. Clip labels are not treated as chronological cues, and place correspondence is assessed using stable structures such as road geometry, building shapes, and fixed installations.

EgoGears-Single review focuses on local visual facts, including objects, spatial relations, camera motion, event order, and environmental changes. Reviewers document the validity of questions and options against the footage.

\paragraph{Post-review checks and repair.}
Following human review, the video model reassesses options without the proposed answer, and the text-only model repeats shortcut screening. Adjudication combines available human judgments, visual verification, text-only results, and proposed answers to retain, revise, repair, or remove questions. Repairs address answer errors, ambiguous wording, incomplete option sets, or insufficient evidence. Repaired questions undergo renewed visual verification and text-only screening.

\paragraph{Release audit.}
Of the $1{,}526$ EgoGears-Multi review items, $706$ are confirmed, $504$ pass repair, $277$ receive answer changes through adjudication, and $39$ are removed, yielding $1{,}487$ released questions.

For EgoGears-Single, $567$ of the $698$ reviewed candidates are retained: $562$ have complete human annotations, while $5$ are retained through video-based verification without complete human annotations. The audit records $120$ answer modifications, $3$ question modifications, and $3$ option modifications. These modifications correct errors in the model-generated candidates; each modified question was re-verified against the footage before release, so the rates describe the candidate pool rather than the released labels.

\subsection{Reproducibility and Validity Checks}
\label{sec:reproducibility}

We independently rescore every per-question result of the 29 reported single-video configurations using the strict parser and recover the originally reported scores exactly. The dataset and prompt are locked by hashes in each run configuration. We also repeat Qwen3-VL-8B on a separate cluster: the direct-answer run obtains 40.9\% versus 40.7\% in the original environment, while the reasoning run obtains 51.0\% versus 52.6\%. The latter difference is explained mainly by a higher residual truncation count (60 versus 38), again emphasizing that serving conditions interact with long reasoning outputs.

Finally, we check whether Gemini's participation in adjudicating 322 multi-video questions explains its lead. Within matched question units, this contributes an estimated 1.5 points to overall Gemini accuracy, far smaller than the 22.1-point gap between Gemini-3.8-Flash and the strongest open model. Table~\ref{tab:human445} reports all configurations on the 445 questions whose reference answers human reviewers confirmed (65 of them after option repair). On this subset, the Gemini models average 80.9\%, compared with 34.0\% across the 28 open-source configurations; Gemini-3.8-Flash leads the strongest open model by 20.5 points (22.1 on the full set), and the ranking of configurations is nearly unchanged (Spearman $\rho=0.99$). Gemini's lead is therefore unlikely to be explained by its role in adjudication. The Gemini models receive the same visual input, prompt, and option shuffling as the open-source models.

\begin{table}[!htb]
\centering
\caption{\textbf{Multi-video exact accuracy on human-confirmed questions} (\%). All: the full set ($N=1{,}487$, as in Table~\ref{tab:multi_results}). Human: the $445$ questions whose reference answers the reviewers confirmed ($65$ after option repair), with $95\%$ bootstrap confidence intervals.}
\label{tab:human445}
\begingroup
\fontsize{7.2}{8.2}\selectfont
\setlength{\tabcolsep}{5pt}
\renewcommand{\arraystretch}{1.1}
\begin{tabular}{@{}lcrrc@{}}
\toprule
Model & Mode & All & Human & 95\% CI \\
\midrule
\rowcolor{egGemini}\multicolumn{5}{@{}l}{\textit{Gemini (API)}}\\
\rowcolor{egGemini}3.8-Flash & API & 78.3 & 90.6 & [87.6, 93.5] \\
\rowcolor{egGemini}3.1-Pro & API & 66.5 & 81.3 & [77.5, 84.9] \\
\rowcolor{egGemini}2.5-Pro & API & 56.4 & 70.8 & [66.5, 74.6] \\
\addlinespace[2pt]
\rowcolor{egQwen35}\multicolumn{5}{@{}l}{\textit{Qwen3.5}}\\
\rowcolor{egQwen35}122B-A10B & T & 37.8 & 49.0 & [44.3, 53.5] \\
\rowcolor{egQwen35} & D & 16.3 & 24.5 & [20.4, 28.5] \\
\rowcolor{egQwen35}35B-A3B & T & 37.9 & 53.3 & [48.1, 58.0] \\
\rowcolor{egQwen35} & D & 15.1 & 24.5 & [20.4, 28.8] \\
\rowcolor{egQwen35}27B & T & 40.8 & 51.2 & [46.5, 55.5] \\
\rowcolor{egQwen35} & D & 16.0 & 22.0 & [18.2, 26.1] \\
\rowcolor{egQwen35}9B & T & 21.3 & 32.4 & [28.1, 36.9] \\
\rowcolor{egQwen35} & D & 10.9 & 20.4 & [16.6, 24.3] \\
\addlinespace[2pt]
\rowcolor{egQwen3VL}\multicolumn{5}{@{}l}{\textit{Qwen3-VL}}\\
\rowcolor{egQwen3VL}235B-A22B & T & 46.0 & 61.3 & [57.1, 66.3] \\
\rowcolor{egQwen3VL} & D & 28.6 & 40.9 & [36.6, 45.4] \\
\rowcolor{egQwen3VL}30B-A3B & T & 26.2 & 39.1 & [34.6, 43.6] \\
\rowcolor{egQwen3VL}8B & D & 16.9 & 25.2 & [21.3, 29.4] \\
\addlinespace[2pt]
\rowcolor{egQwen25VL}\multicolumn{5}{@{}l}{\textit{Qwen2.5-VL}}\\
\rowcolor{egQwen25VL}72B & D & 26.4 & 38.0 & [33.5, 42.2] \\
\rowcolor{egQwen25VL}7B & D & 9.8 & 13.9 & [10.8, 17.1] \\
\addlinespace[2pt]
\rowcolor{egGemma}\multicolumn{5}{@{}l}{\textit{Gemma}}\\
\rowcolor{egGemma}4-31B & T & 56.2 & 70.1 & [65.8, 74.2] \\
\rowcolor{egGemma} & D & 33.2 & 47.9 & [43.4, 52.4] \\
\rowcolor{egGemma}4-26B-A4B & D & 22.0 & 31.2 & [27.0, 35.3] \\
\rowcolor{egGemma}3n-E4B & D & 9.0 & 11.2 & [8.3, 14.2] \\
\addlinespace[2pt]
\rowcolor{egIntern}\multicolumn{5}{@{}l}{\textit{InternVL3.5}}\\
\rowcolor{egIntern}241B-A28B & D & 24.2 & 32.1 & [27.9, 36.6] \\
\rowcolor{egIntern}38B & D & 18.1 & 24.9 & [20.9, 28.8] \\
\rowcolor{egIntern}30B-A3B & D & 13.4 & 21.6 & [17.8, 25.4] \\
\rowcolor{egIntern}20B-A4B & D & 13.9 & 23.8 & [20.0, 27.6] \\
\rowcolor{egIntern}14B & D & 16.5 & 24.9 & [20.9, 28.8] \\
\rowcolor{egIntern}8B & D & 10.4 & 15.1 & [11.7, 18.4] \\
\addlinespace[2pt]
\rowcolor{egGLM}\multicolumn{5}{@{}l}{\textit{GLM}}\\
\rowcolor{egGLM}4.6V & T & 39.9 & 54.2 & [49.9, 58.7] \\
\rowcolor{egGLM} & D & 36.9 & 51.5 & [46.7, 56.2] \\
\rowcolor{egGLM}4.1V-9B & T & 13.4 & 18.4 & [15.1, 21.8] \\
\addlinespace[2pt]
\rowcolor{egKimi}\multicolumn{5}{@{}l}{\textit{Kimi}}\\
\rowcolor{egKimi}VL-A3B-2506 & T & 19.5 & 28.8 & [24.7, 33.0] \\
\bottomrule
\end{tabular}
\endgroup
\end{table}

\begin{figure*}[p]
    \centering
    \includegraphics[width=\textwidth]{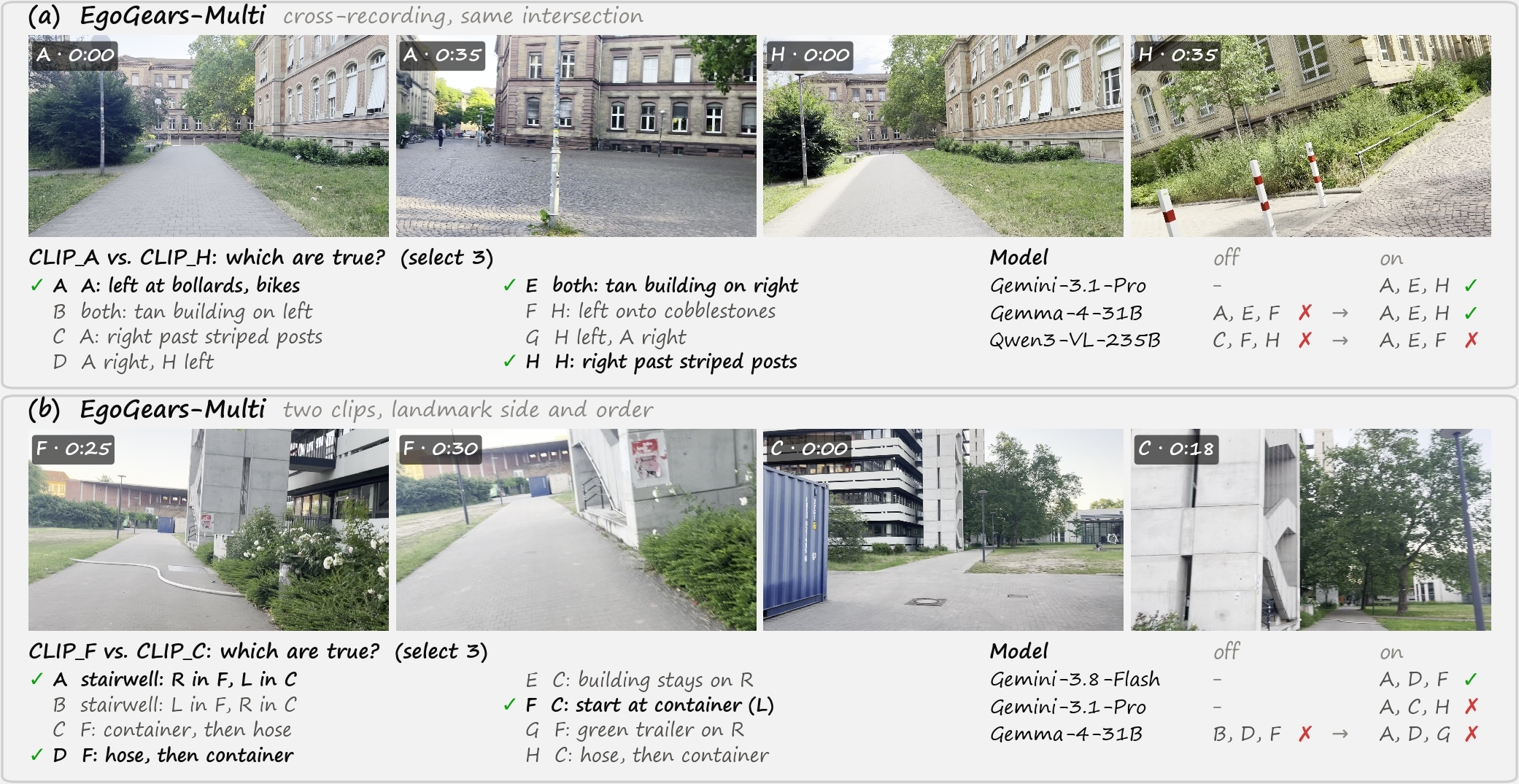}
    \caption{Cross-recording side and order reasoning. The same physical structures must be aligned across observations before direction-dependent relations can be reversed and combined consistently.}
    \label{fig:app_sides_order}
\end{figure*}

Cross-recording comparison adds viewpoint reversal and correspondence uncertainty. In Figure~\ref{fig:app_sides_order}(a), the recordings depict the same intersection, but the correct answer requires reversing the travel-direction-dependent relation while preserving the building identity. Gemma-4-31B is corrected, whereas Qwen3-VL-235B remains inconsistent after reasoning. Panel (b) combines landmark side with within-clip order; only the strongest displayed configuration produces the exact set. These examples show that recognizing shared visual content is insufficient: all selected propositions must be expressed in one consistent coordinate and temporal frame.

\begin{figure*}[p]
    \centering
    \includegraphics[width=\textwidth]{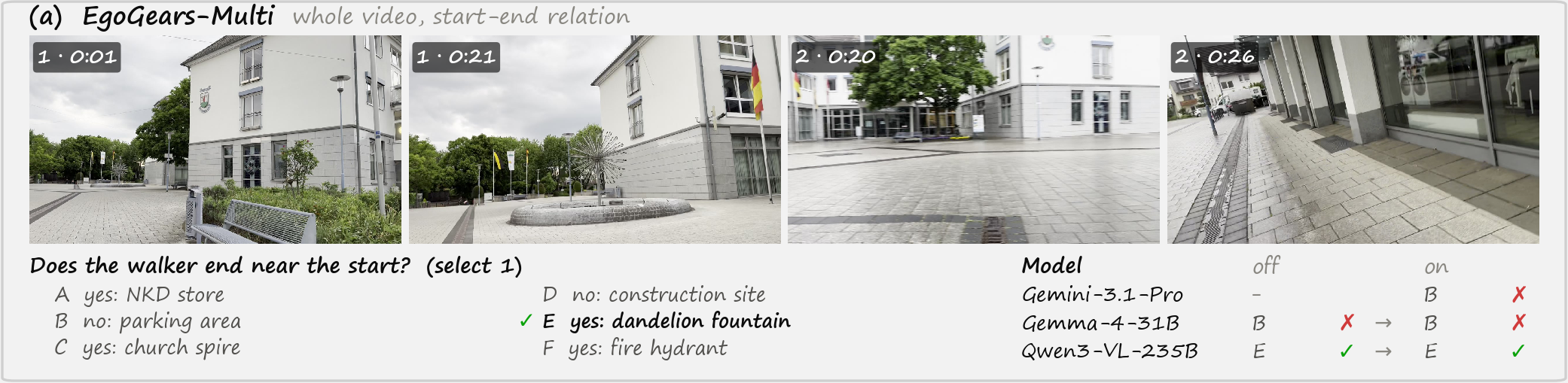}
    \caption{Whole-video start--end correspondence. The models must match distant route phases through a shared landmark; reasoning leaves both the correct and incorrect model predictions unchanged.}
    \label{fig:app_start_end}
\end{figure*}

Figure~\ref{fig:app_start_end} presents a complementary whole-video failure. Determining whether the endpoint returns near the start requires linking distant observations and selecting the landmark that establishes the match. Qwen3-VL-235B succeeds in both modes, while Gemma-4-31B and Gemini-3.1-Pro select the same incorrect alternative. The stable model-specific outcomes suggest that additional reasoning is useful only after the start and end observations have been encoded as comparable place representations.

\subsection{Additional Computation Can Rescue or Destabilize Predictions}

\begin{figure*}[p]
    \centering
    \includegraphics[width=\textwidth]{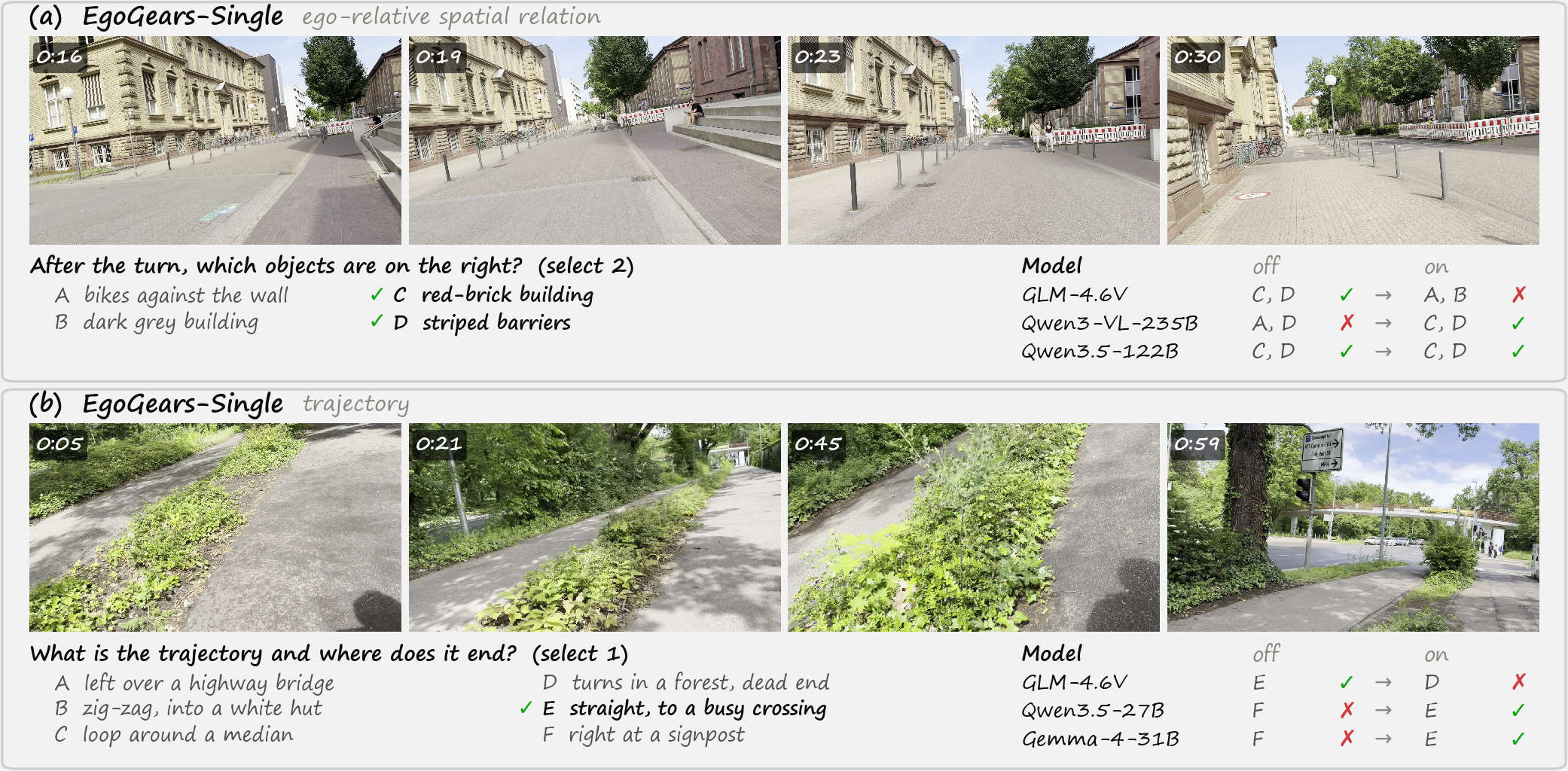}
    \caption{Single-video cases where reasoning has opposite effects across models. Correct direct answers may be destabilized even when another model is rescued on the same question.}
    \label{fig:app_hurts_single}
\end{figure*}

\begin{figure*}[p]
    \centering
    \includegraphics[width=\textwidth]{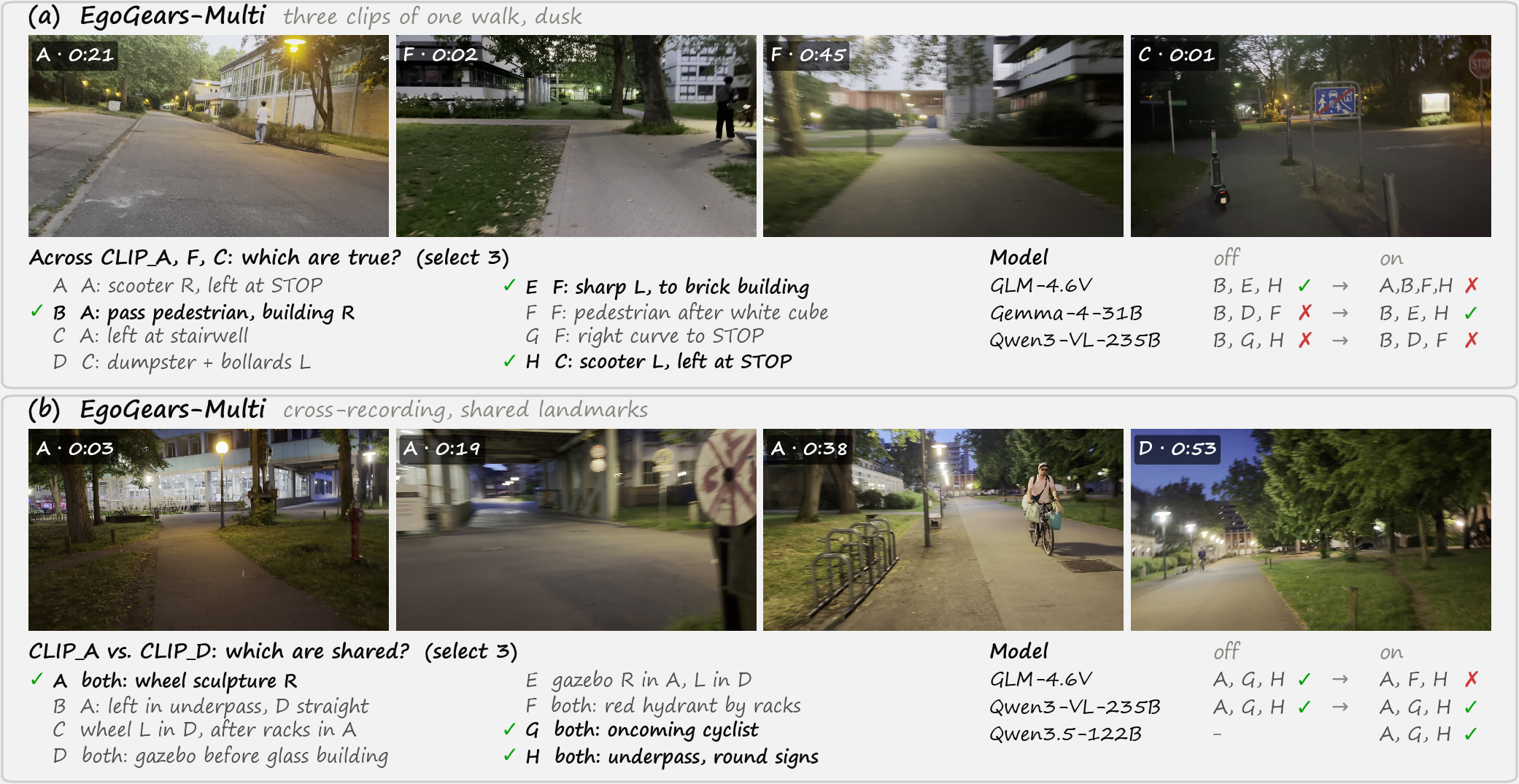}
    \caption{Multi-video regressions and corrections under reasoning. The intervention can improve evidence binding for one model while inducing an inconsistent or wrong-cardinality answer in another.}
    \label{fig:app_hurts_multi}
\end{figure*}

\begin{figure*}[p]
    \centering
    \includegraphics[width=\textwidth]{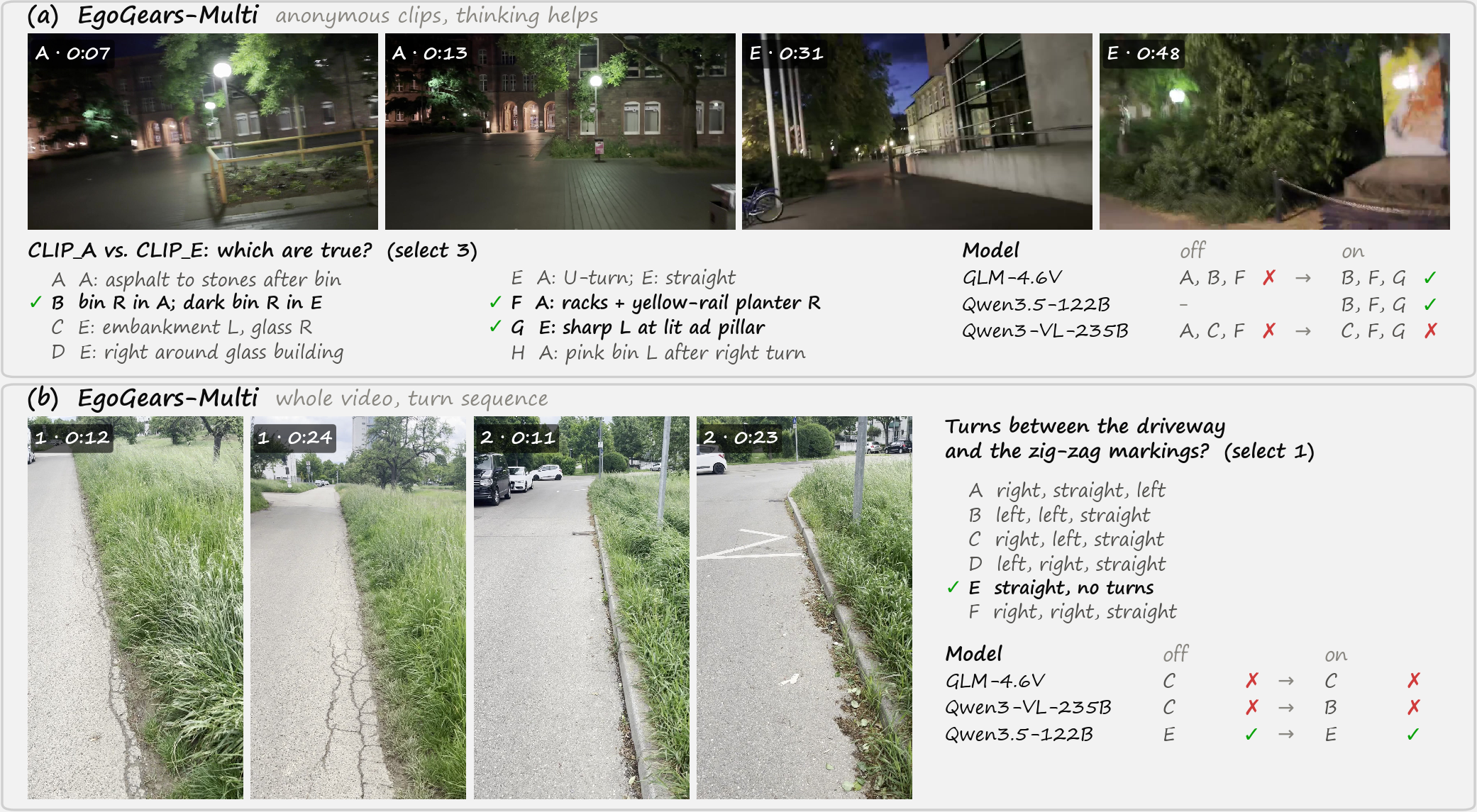}
    \caption{Reasoning rescue versus persistent failure in multi-video QA. Panel (a) contrasts successful and unsuccessful cross-clip evidence integration; panel (b) shows that added computation does not repair an incorrectly inferred turn sequence.}
    \label{fig:app_rescue}
\end{figure*}

Figures~\ref{fig:app_hurts_single}--\ref{fig:app_rescue} compare cases in which reasoning helps, has no effect, or damages an initially correct answer. Figure~\ref{fig:app_hurts_single} is especially instructive because each panel contains both directions of change. In the spatial example, reasoning corrects Qwen3-VL-235B but causes GLM-4.6V to replace the correct set with two unsupported objects; Qwen3.5-122B remains correct. In the trajectory example, Qwen3.5-27B and Gemma-4-31B are rescued, whereas GLM-4.6V changes from the correct endpoint description to an incorrect dead-end interpretation. This pattern is inconsistent with a simple rule that more computation monotonically improves visual reasoning.

The same instability appears in multi-video QA. In Figure~\ref{fig:app_hurts_multi}(a), reasoning rescues Gemma-4-31B but leads GLM-4.6V away from an exact direct answer and does not resolve Qwen3-VL-235B. Panel (b) shows a second GLM regression alongside a stable correct Qwen prediction. Figure~\ref{fig:app_rescue}(a) again contrasts a successful GLM correction with an unresolved Qwen3-VL error. Its panel (b) is a limiting case: two models fail in both modes on an apparently simple ``no turns'' relation, while Qwen3.5-122B is correct in both. Here, reasoning cannot compensate for an incorrect segmentation of route events.

Across these examples, explicit reasoning behaves primarily as an evidence-selection and consistency mechanism, not as a guaranteed source of new visual information. It helps when the required facts are present, but a direct answer binds them to the wrong option, side, or clip. It fails when motion evidence is missing, place correspondence is not established, or the trace introduces an unsupported reinterpretation. This explains why the aggregate gain can coexist with substantial per-instance regressions.

\clearpage

\end{document}